\documentclass[final]{amsart}
\usepackage[utf8]{inputenc} % set input encoding (not needed with XeLaTeX)

\usepackage{graphicx} % support the \includegraphics command and options
\usepackage[foot]{amsaddr} %package that puts the authors addresses on the title page as a footnote (when using AMSART document class only).
\usepackage{verbatim} % adds environment for commenting out blocks of text & for better verbatim
\usepackage{amssymb}
\usepackage{amsthm}
\usepackage{amsmath}
\usepackage{amsfonts}
\usepackage{latexsym}
\usepackage{ulem}
\usepackage{mathtools} % Mathematical tools to use with amsmath http://ftp.math.purdue.edu/mirrors/ctan.org/macros/latex/contrib/mathtools/mathtools.pdf
\usepackage{graphics}
\usepackage{diagbox}
\usepackage{subfigure}
\usepackage{float} %Improved interface for floating objects
\usepackage{enumitem} % http://mirror.las.iastate.edu/tex-archive/macros/latex/contrib/enumitem/enumitem.pdf
\usepackage[english]{babel} 

\usepackage{esint}

\usepackage{calrsfs}
\usepackage{multirow}

\usepackage{hyperref}

\usepackage{dsfont}
\usepackage{xcolor}
\theoremstyle{plain}
\newtheorem{theorem}{Theorem}
\newtheorem{lemma}[theorem]{Lemma}

\newtheorem{proposition}[theorem]{Proposition}

\newtheorem*{remark*}{Remark}

\numberwithin{equation}{section}
\numberwithin{theorem}{section}

\allowdisplaybreaks

\newcommand{\eqdef }{\overset{\mbox{\tiny{def}}}{=}}

\newcommand{\rone}{\mathbb{R}}

\newcommand{\norm}[1]{\left\lVert{#1}\right\rVert}
\newcommand{\reviewerX}[1]{#1}
\newcommand{\reviewerY}[1]{#1}
\newcommand{\reviewerZ}[1]{#1}
\newcommand{\reviewerTWO}[1]{#1}

\title[FourierSpecNet for Solving the Boltzmann Equation]{FourierSpecNet: Neural Collision Operator Approximation Inspired by the Fourier Spectral Method for Solving the Boltzmann Equation}

\author[J. Y. Lee]{Jae Yong Lee$^1$}
\address{$^1$Department of AI, Chung-Ang University, Seoul 02455, Republic of Korea. \href{mailto:jaeyong@cau.ac.kr}{jaeyong@cau.ac.kr} }

\author[G. J. Jung]{Gwang Jae Jung$^{2, \ddagger}$}
\address{$^2$Department of Mathematics, Pohang University of Science and Technology (POSTECH), Pohang 37673, Republic of Korea.  \href{mailto:gjjung@postech.ac.kr}{gjjung@postech.ac.kr}}

\author[B. C. Lim]{Byung Chan Lim$^{3, \ddagger}$}
\address{$^3$Department of Mathematics, Pohang University of Science and Technology (POSTECH), Pohang 37673, Republic of Korea.  \href{mailto:limbc@postech.ac.kr}{limbc@postech.ac.kr}}

\author[H. J. Hwang]{Hyung Ju Hwang$^{4,\dagger}$}
\address{$^4$Department of Mathematics, Pohang University of Science and Technology (POSTECH), Pohang 37673, Republic of Korea. \href{mailto:hjhwang@postech.ac.kr}{hjhwang@postech.ac.kr}}
\address{$^\ddagger$Equal contribution.}
\address{$^\dagger$Corresponding author.}

\AtBeginDocument{%
   \def\MR#1{}
}

\begin{document}

%\keywords{collisional Kinetic Theory}
%\subjclass[2000]{}

%\date{\today}
%\date{\today; \Red{(DRAFT)}}
% \date{May 8, 2019}
% \date{}

\let\thefootnote\relax
\addtocounter{footnote}{-1}\let\thefootnote\svthefootnote

\begin{abstract}
The Boltzmann equation, a fundamental model in kinetic theory, describes the evolution of particle distribution functions through a nonlinear, high-dimensional collision operator. However, its numerical solution remains computationally demanding, particularly for inelastic collisions and high-dimensional velocity domains. In this work, we propose the Fourier Neural Spectral Network (FourierSpecNet), a hybrid framework that integrates the Fourier spectral method with deep learning to approximate the collision operator in Fourier space efficiently. FourierSpecNet achieves resolution-invariant learning and supports zero-shot super-resolution, enabling accurate predictions at unseen resolutions without retraining. Beyond empirical validation, we establish a consistency result showing that the trained operator converges to the spectral solution as the discretization is refined. We evaluate our method on several benchmark cases, including Maxwellian and hard-sphere molecular models, as well as inelastic collision scenarios. The results demonstrate that FourierSpecNet offers competitive accuracy while significantly reducing computational cost compared to traditional spectral solvers. Our approach provides a robust and scalable alternative for solving the Boltzmann equation across both elastic and inelastic regimes.
\end{abstract}

% set the depth for the table of contents (0-2)
\setcounter{tocdepth}{2}
%\chapter is level 0
%\section is level 1
%\subsection is level 2
%\subsubsection is level 3
%\paragraph is level 4
%\subparagraph is level 5

\maketitle
\tableofcontents

\thispagestyle{empty}

\section{Introduction}
\subsection{Motivation and main results}
%% Boltzmann equation
The Boltzmann equation~\cite{MR0158708} is a fundamental mathematical framework for modeling the evolution of particle systems in rarefied gases. It captures the statistical behavior of particles undergoing binary collisions, making it indispensable in fields like fluid dynamics, aerospace engineering, and plasma physics~\cite{Vil02}. The collision operator, a high-dimensional non-linear integral term that governs particle interactions in the Boltzmann equation, presents a significant computational challenge in numerical simulations. Solving the Boltzmann equation efficiently is a longstanding problem in computational mathematics due to the vast computational resources required to approximate its solutions accurately.

%% Various methods
Various numerical methods have been developed, broadly classified into probabilistic and deterministic approaches for solving kinetic equations. Probabilistic methods, such as the direct simulation Monte Carlo (DSMC)~\cite{bird1994molecular} technique, rely on random sampling to model particle interactions. While computationally flexible and widely used in simulation, these methods often suffer from statistical noise and slow convergence, especially when high accuracy is needed. On the other hand, deterministic methods aim to directly approximate the collision operator using structured mathematical frameworks. Among these, the Fourier spectral method~\cite{MR1407541} has emerged as a particularly powerful approach. By leveraging the Fourier transform, the Fourier spectral method achieves spectral accuracy, providing highly precise approximations of the collision operator. Additionally, the Fourier spectral method is renowned for its stability when solving the Boltzmann equation~\cite{MR1756425,MR1770438}. However, its traditional implementation has $O(N^{2d})$ computational complexity, where $N$ is the resolution of the discretization of the velocity space and $d$ is the dimension of the velocity space.
This high computational cost makes it impractical for large-scale or high-dimensional problems. To address this limitation, many studies developed fast spectral methods to reduce the computational cost $O(N^{2d})$ of approximating the Boltzmann equation as $O(N^d\log N)$~\cite{pareschi2003computational}. They use the fast Fourier transform (FFT) to compute the weighted convolution in the discrete velocity space. These methods compute the weighted convolution in a discrete velocity space more efficiently, enabling the application of spectral techniques to problems that require solutions of high resolution. Despite these advances, deterministic methods still face scalability issues, particularly for high-dimensional velocity spaces or complex collision kernels.

%% NN methods
In parallel, recent advances in deep learning have introduced a new paradigm for solving partial differential equations (PDEs). Methods like Physics-Informed Neural Networks (PINNs)~\cite{karniadakis2021physics} and operator learning frameworks such as the Fourier Neural Operator (FNO)~\cite{li2021fourier} and deep operator network (DeepONet)~\cite{deeponet} have demonstrated the potential of neural networks to efficiently approximate complex PDE solutions. These methods leverage neural network architectures to generalize across different resolutions and conditions, achieving capabilities like zero-shot super-resolution~\cite{li2021fourier}. Recently, there have been many trials to approximate the solution of kinetic equations using recent physics-informed machine learning techniques~\cite{MR4116803,MR4313375,miller2021encoder,MR4554720,MR4280255,MR4609518,MR4887484,lee2024structure}. Despite these advancements, many of these approaches primarily focus on designing deep learning architectures to approximate the PDE solutions, rather than incorporating the strengths of established numerical solvers. Challenges such as ensuring theoretical convergence and scalability in high-dimensional problems remain, whereas traditional numerical methods, although well-established, often struggle to balance accuracy and computational efficiency, especially in complex or high-resolution regimes.

%% Combine spectral method with nn => pros
Motivated by these challenges, we propose the Fourier Neural Spectral Network (FourierSpecNet), a novel hybrid framework that combines the Fourier spectral method with deep learning.
% key:not_replacing_precomputation
\reviewerY{FourierSpecNet computes the collision term of the given input distributions using the structure of the fast spectral method, while the parameters in the fast spectral method are computed using deep learning. 
Although both FourierSpecNet and the fast spectral method rely on precomputed parameters that are repeatedly invoked during simulation, the computational efficiency of FourierSpecNet does not stem from avoiding this precomputation stage. 
Instead, FourierSpecNet achieves higher efficiency through architectural modifications that fundamentally reduce the dependence of the separable rank on the input resolution, thereby decreasing overall computational complexity.}
%%%%%%%%%%%%%%%%%%%%%%%%%
% 아래의 문장은 삭제해도 괜찮을 것 같은데, 어떻게 생각하시는지 궁금합니다.
FourierSpecNet retains the spectral accuracy and stability of traditional numerical methods while overcoming their computational limitations.
%%%%%%%%%%%%%%%%%%%%%%%%%
% By embedding neural network parameters directly in Fourier space, it introduces resolution invariance and zero-shot super-resolution capabilities, enabling high-resolution predictions even from low-resolution training data.
\reviewerX{By embedding neural network parameters directly in a fixed truncated Fourier domain, FourierSpecNet achieves resolution invariance, since its parameter count depends only on the truncated spectral bandwidth and remains independent of the inference grid. 
This structural property, rather than a generic feature of neural networks, naturally enables zero-shot super-resolution, allowing high-resolution inference from models trained on low-resolution data. 
This mechanism significantly reduces computational cost while maintaining accuracy, making FourierSpecNet suitable for applications requiring dynamic resolution adaptation, such as multi-scale modeling in fluid dynamics and high-dimensional kinetic simulations.}

Furthermore, FourierSpecNet can approximate the Boltzmann collision operator for a wide range of collision kernels, including Maxwellian molecules, hard-sphere molecules, and inelastic cases, using data-driven learning. By training on diverse input distributions and their corresponding collision terms, the model generalizes effectively to unseen scenarios without additional computational effort. GPU acceleration further enhances computational efficiency, allowing batch-wise inference that dramatically reduces computational time compared to classical Fourier spectral methods.

An important innovation of FourierSpecNet is its resolution-invariance property. \reviewerX{The property arises from the spectral parameterization itself: since the learnable coefficients are defined only on a fixed truncated Fourier space, the total number of parameters remains constant regardless of the discretization grid size. 
This structural formulation explains why FourierSpecNet can perform reliable high-resolution inference without retraining, distinguishing it from the traditional spectral method.} This capability was validated through numerical experiments, where the model demonstrated robust performance and scalability across various resolutions, achieving high accuracy and efficiency while preserving physical properties such as mass, momentum, and energy. In addition to empirical performance, we also provide a theoretical consistency guarantee \reviewerX{in Proposition \ref{prop:consistency_gjjung}} showing that the approximation error of FourierSpecNet is bounded by the spectral truncation error, ensuring that the trained operator converges to the true Boltzmann collision operator as the resolution increases. Our main contribution is as follows.
\begin{itemize}
\item \textbf{Hybrid Framework}: We present a novel Boltzmann equation solver that combines deep learning with the fast spectral method, a widely used technique for approximating solutions to the Boltzmann equation. To the best of our knowledge, this method represents the first deep learning framework based on the fast spectral method for solving the Boltzmann equation.
\item \textbf{Resolution Invariance and Super-resolution}: Utilizing neural network parameters in Fourier space, FourierSpecNet is resolution-invariant, enabling zero-shot super-resolution. This allows high-resolution inference from low-resolution training data, significantly reducing computational costs. 
\item \textbf{Versatility}: The proposed framework supports diverse collision kernels, including Maxwellian molecules, hard-sphere molecules, and inelastic cases, demonstrating robust generalization across various kinetic scenarios.
\item \textbf{Computational Efficiency and Theoretical evidence}: By leveraging GPU acceleration, FourierSpecNet achieves rapid batch-wise inference of the collision operator, reducing computational time while maintaining certain accuracy in preserving physical properties like mass, momentum, and energy. The approximation error for the collision operator is also shown to be theoretically consistent with the spectral truncation error.
\end{itemize}

\subsection{Boltzmann equation}
In kinetic theory, the function $f(t,\boldsymbol{x},\boldsymbol{v})$ denotes the density of particles moving with velocity $\boldsymbol{v}$ in position $\boldsymbol{x}$ at time $t$.
The full Boltzmann equation is expressed as
\begin{equation}
    % \frac{\partial f}{\partial t}+\boldsymbol{v}\cdot \nabla_{\boldsymbol{x}}f = \frac{1}{K_n}Q(f,f),
    \frac{\partial f}{\partial t}+\boldsymbol{v}\cdot \nabla_{\boldsymbol{x}}f = Q(f,f),
\end{equation}
where $\boldsymbol{x},\boldsymbol{v}\in\rone^d$ with $d=2$ or $d=3$, and
% \begin{equation}\label{eq:collision}
%   \begin{split}
%   Q(f,f)&=Q^+(f,f)-L[f]f\\
%     &= \int_{\rone^d}\int_{S^{d-1}} B(|\boldsymbol{v}-\boldsymbol{v}_*|,\cos\theta)f(\boldsymbol{v}')f(\boldsymbol{v}_*')d\omega d\boldsymbol{v}_*\\
%     &\quad-f(\boldsymbol{v})\int_{\rone^d}\int_{S^{d-1}} B(|\boldsymbol{v}-\boldsymbol{v}_*|,\cos\theta)f(\boldsymbol{v}_*)d\omega d\boldsymbol{v}_*
%   \end{split}
% \end{equation}
\begin{equation}\label{eq:collision}
\begin{split}
    Q(f,f)
    &=
    \int_{\mathbb{R}^d} \int_{S^{d-1}}{
        B(|\boldsymbol{v}-\boldsymbol{v}_*|, \cos\theta)
        \left(
            f(\boldsymbol{v}')f(\boldsymbol{v}_*')
            -
            f(\boldsymbol{v}) f(\boldsymbol{v_*})
        \right)
    }
    d\omega d\boldsymbol{v}_*.
    % &\quad-f(\boldsymbol{v})\int_{\rone^d}\int_{S^{d-1}} B(|\boldsymbol{v}-\boldsymbol{v}_*|,\cos\theta)f(\boldsymbol{v}_*)d\omega d\boldsymbol{v}_*
\end{split}
\end{equation}
Here, $\boldsymbol{v}'$ and $\boldsymbol{v}_*'$ are defined as
% \begin{align}
% \begin{split}
%     \boldsymbol{v}'
%     &=
%     \frac{1}{2}(\boldsymbol{v}+\boldsymbol{v}_*+|\boldsymbol{v}-\boldsymbol{v}_*|\omega),
%     \\
%     \boldsymbol{v}_*'
%     &=
%     \frac{1}{2}(\boldsymbol{v}+\boldsymbol{v}_*-|\boldsymbol{v}-\boldsymbol{v}_*|\omega).
% \end{split}
% \end{align}
\begin{equation}
    \boldsymbol{v}'=\frac{1}{2}(\boldsymbol{v}+\boldsymbol{v}_*+|\boldsymbol{v}-\boldsymbol{v}_*|\omega),
    \quad\quad
    \boldsymbol{v}_*'=\frac{1}{2}(\boldsymbol{v}+\boldsymbol{v}_*-|\boldsymbol{v}-\boldsymbol{v}_*|\omega).
\end{equation}
For the variable hard sphere (VHS) model, we can consider the kernel $B(|\boldsymbol{v}-\boldsymbol{v}_*|,\cos\theta)$ as
$$B(|\boldsymbol{v}-\boldsymbol{v}_*|,\cos\theta)=C|\boldsymbol{v}-\boldsymbol{v}_*|^\alpha.$$
The Maxwellian gas corresponds to $\alpha=0$, and the hard sphere gas corresponds to $\alpha=1$. Additionally, there exist other collision kernels for different cases. More generally, scenarios involving energy loss during collisions can also be considered. The elasticity parameter $e$, also known as the restitution coefficient, quantifies the energy dissipation that occurs during collisions. In the inelastic case, the post-collision velocities $\boldsymbol{v}^\prime$ and $\boldsymbol{v}_*^\prime$ in \eqref{eq:collision} are defined as
% \begin{equation}
%     \boldsymbol{v}'
%     =
%     \frac{1}{2}(\boldsymbol{v}+\boldsymbol{v}_*+\frac{1+e}{2}|\boldsymbol{v}-\boldsymbol{v}_*|\omega)
% \end{equation}
% and 
% \begin{equation}
%     \boldsymbol{v}_*'
%     =
%     \frac{1}{2}(\boldsymbol{v}+\boldsymbol{v}_*-\frac{1+e}{2}|\boldsymbol{v}-\boldsymbol{v}_*|\omega).
% \end{equation}
\begin{align}
\label{post-collision velocity}
\begin{split}
    \boldsymbol{v}
    &=
    \frac{\boldsymbol{v} + \boldsymbol{v}_*}{2}
    +
    \frac{1-e}{4}(\boldsymbol{v} - \boldsymbol{v}_*)
    +
    \frac{1+e}{4} |\boldsymbol{v} - \boldsymbol{v}_*| \omega,
    \\
    \boldsymbol{v}'
    &=
    \frac{\boldsymbol{v} + \boldsymbol{v}_*}{2}
    -
    \frac{1-e}{4}(\boldsymbol{v} - \boldsymbol{v}_*)
    -
    \frac{1+e}{4} |\boldsymbol{v} - \boldsymbol{v}_*| \omega.
\end{split}
\end{align}
% The coefficient of restitution $e$ characterizes the degree of inelasticity in collisions. The special case $e = 1$ corresponds to the elastic Boltzmann equation, where no energy is dissipated during collisions. The case $0 \leq e < 1$ is the inelastic Boltzmann.
The coefficient of restitution $e$ characterizes the degree of elasticity of a collision; collisions are elastic when $e=1$, and inelastic when $0 \leq e < 1$.
The full Boltzmann equation can be solved by solving the following two steps:
\begin{align*}
    &\text{Transport stage:}\quad\frac{\partial f}{\partial t} + \boldsymbol{v}\cdot \nabla_x f = 0
    \\
    &\text{Collision stage:}\quad\frac{\partial f}{\partial t} = Q(f,f)
\end{align*}
In this study, we focus on the space-homogeneous Boltzmann equation for the collision stage. We try to approximate the solution $f(t,\boldsymbol{v})$ to the following equation:
\begin{equation*}
    \frac{\partial f}{\partial t}=Q(f,f).
\end{equation*}
The challenging aspect of the Boltzmann equation lies in the operator $Q(f,f)$ that varies with each $f$. Once $Q(f,f)$ is well approximated, solutions $f$ to the equation over time can be obtained using methods such as Euler or Runge-Kutta, e.g. using the approximated operator $Q(f, f)$, we can get the distribution at next time $t_{n+1}$ as follows:
\begin{equation}
  f(t_{n+1},\boldsymbol{v})=f(t_n,\boldsymbol{v})+\Delta t \, Q(f, f)(t_n, \boldsymbol{v}).
\end{equation}

\subsection{Fundamental properties of physical quantities and the equilibrium state}\label{properties}
It is well known that the collision operator $Q(f,f)$ preserves mass, momentum and energy during the evolution of particles, i.e.,
\begin{equation}
    \int_{\mathbb{R}^d} Q(f,f)\phi(v) d\boldsymbol{v}=0,
\end{equation}
where $\phi(v):=(1,\boldsymbol{v},|\boldsymbol{v}|^2)$. The collision operator also satisfies $\mathit{H}$-theorem
\begin{equation}
    \int_{\mathbb{R}^d} Q(f,f) \ln f d\boldsymbol{v}\leq0
\end{equation}
which implies the equality $Q(f,f)=0$ if and only if the distribution $f(t,\boldsymbol{x},\boldsymbol{v})$ is a form of Maxwellian $M[f]$ where
\begin{equation}
    M[f](\boldsymbol{v}):=\frac{\rho}{(2\pi T)^{d/2}}\exp\left(-\frac{|\boldsymbol{v}-\boldsymbol{u}|^2}{2T}\right)
\end{equation}
with the macroscopic density $\rho$, mean velocity $u$, and temperature $T$ defined as
\begin{equation}\label{eq:physics_quantity}
    \rho:=\int_{\mathbb{R}^d} f d\boldsymbol{v}, \; \boldsymbol{u}:=\frac{1}{\rho}\int_{\mathbb{R}^d} f\boldsymbol{v} d\boldsymbol{v}, \; T:=\frac{1}{d\rho}\int_{\mathbb{R}^d} f|\boldsymbol{u}-\boldsymbol{v}|^2 d\boldsymbol{v}.
\end{equation}
The Boltzmann equation conserves the physical quantities, such as mass, momentum and kinetic energy defined as
\begin{equation}
    \text{Mass}(t):=\rho, \; \text{Mom}(t):=\rho\boldsymbol{u}, \; \text{KE}(t):=\rho|\boldsymbol{u}|^2+\rho dT.
\end{equation}

\subsection{Related works}
% Boltzmann theory
Many theoretical studies on the Boltzmann equation have a long history. The existence and uniqueness of a solution in the space-homogeneous case was first established by Carleman~\cite{carleman1933theorie} in the 1930s, and subsequently developed through works such as~\cite{ukai1974existence,MR1014927,MR760333}. We refer to~\cite{Vil02} for a comprehensive overview of the theoretical analysis of the Boltzmann equation.

% Numerical methods
Numerical methods for simulating the Boltzmann equation largely focus on approximating the high-dimensional, nonlinear structure of collision operator in~\eqref{eq:collision}. Stochastic approaches such as direct simulation Monte Carlo (DSMC)~\cite{bird1994molecular,nanbu1980direct,nanbu1983interrelations} have been developed to address the curse of dimensionality. However, they are computationally expensive to converge, especially for high accuracy. Deterministic methods offer an alternative, including the discrete velocity method~\cite{goldstein1989investigations,MR1322351,wagner1995approximation,MR1380030,MR3100773} and finite-difference method~\cite{sone1989temperature,kosuge2001shock}. See~\cite{MR3202241,degond2004modeling} for more details and other numerical techniques. Among deterministic solvers, the Fourier spectral method~\cite{MR1407541,MR1728639}, which employs the fast Fourier transform (FFT), has demonstrated high accuracy. Pareschi and Russo~\cite{MR1756425} proposed an algorithm with spectral accuracy and computational complexity $O(N^{2d})$ in velocity space, while Filbet and Russo~\cite{MR1973198} extended this framework to spatially inhomogeneous problems. Later, Mouhot and Pareschi~\cite{MR2240637} introduced a faster algorithm using Carleman representation, reducing the complexity to $O(MN^d\log N)$ for hard-sphere molecules. Gamba et al.~\cite{MR3687853} further generalized this to arbitrary collision kernels with complexity $O(MN^{d+1}\log N)$. The inelastic Boltzmann collision operator has also been studied via spectral methods in~\cite{MR3920936}.

% Neural network approaches
In recent years, deep learning-based methods have been developed in the numerical PDE community~\cite{karniadakis2021physics}. Physics-Informed Neural Networks (PINNs)~\cite{MR3881695} and their variants~\cite{deeponet,MR4888707,li2021fourier} leverage the universal approximation capabilities of neural networks~\cite{hornik1989multilayer} to simulate PDEs directly. These advancements include the Fourier Neural Operator proposed by Li et al.~\cite{li2021fourier}, which retains the same training network parameters regardless of the discretization of input and output spaces. Notably, this method enables zero-shot super-resolution, allowing models trained on lower resolutions to be directly evaluated at higher resolutions without retraining. 

More recently, there have been growing efforts to apply these approaches to kinetic equations, including the Vlasov-Poisson-Fokker-Planck~\cite{MR4116803,MR4313375}, Landau~\cite{miller2021encoder,MR4554720}, and Boltzmann~\cite{MR4280255,MR4609518} equations. In particular, several works focus on embedding physical structure into neural networks to enhance their accuracy and interpretability. For example, SPINNs~\cite{MR4887484} reduce quadrature costs for the BGK model by leveraging tensor decomposition and Gaussian basis functions. Meanwhile, structure-preserving DeepONets~\cite{lee2024structure} incorporate conservation laws such as mass invariance directly into the network architecture. These methods demonstrate a growing trend of integrating domain knowledge into deep learning for kinetic modeling. However, most of the aforementioned works primarily focus on approximating Boltzmann equation through carefully designed network architectures, without directly leveraging classical numerical solvers. While neural models can generalize across conditions, challenges remain in achieving theoretical convergence and scalability in high-dimensional settings. Motivated by this gap, our proposed Fourier Neural Spectral Network (FourierSpecNet) seeks to bridge the strengths of both worlds.

\subsection{Outline of the paper}
In the following Section \ref{sec:fourier}, we provide an overview of the traditional Fourier spectral method. Section \ref{sec:method} introduces the proposed approach, the Fourier Neural Spectral Network, which combines deep learning with the existing Fourier spectral method and includes a theoretical consistency analysis of the trained operator. Following this, Section \ref{sec:simulation} presents simulation results obtained using this method. The simulations encompass scenarios ranging from Maxwellian molecules to hard spheres, as well as inelastic cases and three-dimensional scenarios. Section \ref{sec:conclusion} offers a summary of our proposed approach.

\section{Fourier Spectral Method}\label{sec:fourier}
In this section, we present the Fourier spectral method, which is one of the most efficient numerical methods for approximating the collision term $Q(f,f)$. 
We assume that $f$ has a compact support $\mathit{B}_S$, the ball of radius $S$ centered at the origin. Although the compact support of $f$ expands as the collision proceeds, we can ensure that it does not exceed a ball $\mathit{B}_R$ with $R=2S$. Additionally, we extend $f$ to a periodic function on the set $[-T,T]^d$ with $T=\frac{3+\sqrt{2}}{2}S$. The choice of $T$ prevents intersections of the regions where $f$ is different from zero. %Under these conditions we can ensure $Q(f,f)$ also has a compact support with a ball $\mathit{B}_{\sqrt{2}R}$, so denote by $Q^R(f)$ the collision term with \textit{cutoff}.

To simplify notation, let us define $\mathit{D}_T = [-T,T]^d$, and assume $k, l, m \in \mathbb{Z}^d$.
% The Fourier Transform of $Q$ also can be expressed as
%%%%%%%%%%%%%%%%%%%%%%%%%%%%%%%%%%%%%%%%%%%%%%%%%%
%%%%%%%%%%%%%%%%%%%%%%%%%%%%%%%%%%%%%%%%%%%%%%%%%%
% Collision operator and spectral method
% \input{__backup_FSM__limbc}
For each positive integer $N$, let $f_N$ be the truncated Fourier series of $f$ of degree less than or equal to $N$, i.e.,
\begin{align}\label{eq:trunk_f}
    f_N(t, \boldsymbol{v})
    =
    \sum_{\| k \|_\infty \leq N} \hat{f}_k(t) e^{i \frac{\pi}{T} k \cdot \boldsymbol{v}}
    \quad
    (\boldsymbol{v} \in D_T),
\end{align}
where
\begin{align}
    \hat{f}_k(t)
    =
    \frac{1}{(2T)^d} \int_{\mathit{D}_T} f(t, \boldsymbol{v}) e^{-i \frac{\pi}{T} k \cdot \boldsymbol{v}} d\boldsymbol{v}.
\end{align}
The Fourier-Galerkin method seeks for the coefficients $\hat{f}_k$ in the Fourier series $f_N$ for $k \in \mathbb{Z}^d$ with $\| k \|_\infty \leq N$ which satisfy
\begin{align}
    \int_{D_T} \left( \frac{\partial f_N}{\partial t} - Q(f_N, f_N) \right) e^{-i \frac{\pi}{T} k \cdot \boldsymbol{v}} = 0,
\end{align}
or equivalently
\begin{align}
    \frac{\partial \hat{f}_k}{\partial t}
    =
    \frac{1}{(2T)^d} \int_{D_T} Q(f_N, f_N) e^{-i \frac{\pi}{T} k \cdot \boldsymbol{v}}\, d\boldsymbol{v}.
\end{align}
Defining $q \coloneqq v_* - v$ and $q^+ \coloneqq \dfrac{q+|q|\omega}{2}$, we have
\begin{equation}\label{eq:post-collision_velocity_with_q}
    \boldsymbol{v}' = \boldsymbol{v} + \dfrac{1+e}{2} q^+,
    \,
    \boldsymbol{v}_*' = \boldsymbol{v}_* - \dfrac{1+e}{2} q^+,
\end{equation}
so plugging (\ref{eq:trunk_f}) and (\ref{eq:post-collision_velocity_with_q}) into (\ref{eq:collision}), we have
% \begin{align}
% \begin{split}
%     Q(f_N, f_N)
%     &=
%     \sum_{\|l\|_\infty, \|m\|_\infty \leq N} \int_{B_0(2S)} \int_{S^{d-1}}
%     \hat{f}_l \hat{f}_m
%     e^{ i \frac{\pi}{T} (l+m) \cdot \boldsymbol{v} }
%     \\
%     &\phantom{  % Phantom begins
%         =
%         \sum_{\|l\|_\infty, \|m\|_\infty \leq N} \int_{B_0(2S)} \int_{S^{d-1}}
%     }           % Phantom ends
%         \left(
%             e^{ i \frac{\pi}{T} \left( m \cdot q + (l-m) \cdot \frac{1+e}{2}q^+ \right) }
%             -
%             e^{ i \frac{\pi}{T} m \cdot q }
%         \right)
%     \, d\omega \, dq.
% \end{split}
% \end{align}
\begin{multline}
    Q(f_N, f_N)=\\
    \sum_{\|l\|_\infty, \|m\|_\infty \leq N} \int_{B_0(2S)} \int_{S^{d-1}}
    \hat{f}_l \hat{f}_m e^{ i \frac{\pi}{T} (l+m) \cdot \boldsymbol{v} }
        \left(
            e^{ i \frac{\pi}{T} \left( m \cdot q + (l-m) \cdot \frac{1+e}{2}q^+ \right) }
            -
            e^{ i \frac{\pi}{T} m \cdot q }
        \right)
    \, d\omega \, dq.
\end{multline}

Therefore, for $k \in \mathbb{Z}^d$ with $\| k \|_\infty \leq N$, the $k$-th Fourier coefficient $\hat{Q}_k(f_N, f_N)$ of $Q(f_N, f_N)$ can be computed as
\begin{equation}
    \hat{Q}_k(f_N,f_N)
    =
    \sum \limits_{\substack{\|l\|_\infty, \|m\|_\infty \leq N \\ l+m=k}}^{} \hat{B}(l,m) \hat{f}_l \hat{f}_m,
\end{equation}
where $\hat{B}(l,m) \coloneqq \hat{G}(l, m) + \hat{g}(m)$ with
\begin{align}\label{eq:fourier_weight}
\begin{split}
    \hat{G}(l,m)
    &\coloneqq
    \int_{B_0(2S)} \int_{S^{d-1}}
        B( |q|, \hat{q}\cdot\omega )
        \exp\left(  i \frac{\pi}{T} q \cdot \left( \frac{1+e}{4}l - \frac{3-e}{4}m \right) \right)
    \\
    &\phantom{  % Phantom begins
        \coloneqq
        \int_{B_0(2S)} \int_{S^{d-1}}
    }           % Phantom ends
            \exp\left( -i \frac{\pi}{T} |q| \omega \cdot \frac{1+e}{4}(l+m) \right)
        \, d\omega \, dq,
\end{split}
\end{align}
and $\hat{g}(m)\coloneqq-\hat{G}(m, m)$.
% Collision operator and spectral method
%%%%%%%%%%%%%%%%%%%%%%%%%%%%%%%%%%%%%%%%%%%%%%%%%%
%%%%%%%%%%%%%%%%%%%%%%%%%%%%%%%%%%%%%%%%%%%%%%%%%%

The total cost of evaluating $\hat{Q}_k$ is $O(N^{2d})$, where $N$ is the number of grid points in the discretized velocity domain. This becomes computationally prohibitive, particularly as the dimension increases.
To address this challenge, many studies approximate $\hat{G}(l,m)$ as a sum of separable functions of $l+m,\, l,\, m$,
% (Correction 2) To address this challenge, many studies approximate $\hat{Q}_k(f,f)$ as a weighted sum of discrete convolutions,
% (Before) To address this challenge, many studies approximate $\hat{G}_(l,m)$ as a weighted sum of discrete convolutions,
enabling compatibility with standard $O(N^d \log N)$ FFT algorithms such as the Cooley–Tukey algorithm.
% Recently, Gamba et al.~\cite{MR3687853} proposed an approximation of $\hat{G}(l,m)$ in the form:
Recently, Gamba et al.~\cite{MR3687853} proposed an approximation of $\hat{G}(l,m)$ for the case of elastic collision, by applying quadrature rules on equation \eqref{eq:fourier_weight}.
Applying an analogous computation in \cite{MR3687853} to the inelastic case, we have
% \begin{align}
% \label{eq:kernel_modes_before_approximation}
% \begin{split}
%     \hat{G}(l,m)
%     &=
%     \int_0^{2S} \int_{S^{d-1}} \int_{S^{d-1}}
%     B(r, \sigma \cdot \omega)
%     e^{ -i \frac{\pi}{2T} r \omega \cdot \frac{1+e}{2}(l+m) }
%     \\
%     &\phantom{  % Phantom begins
%         =
%         \int_0^{2S} \int_{S^{d-1}} \int_{S^{d-1}}
%     }           % Phantom ends
%         e^{  i \frac{\pi}{2T} r \sigma \cdot \frac{1+e}{2}l }
%         e^{ -i \frac{\pi}{2T} r \sigma \cdot \frac{3-e}{2}m }
%     \, d\omega \, d\sigma \, dr,
%     \\
%     &=
%     \int_0^{2S} \int_{S^{d-1}} {
%         F(l+m; r, \sigma)
%         e^{  i \frac{\pi}{2T} r \sigma \cdot \frac{1+e}{2}l }
%         e^{ -i \frac{\pi}{2T} r \sigma \cdot \frac{3-e}{2}m }
%     } \, d\sigma \, dr,
% \end{split}
% \end{align}
\begin{align}\label{eq:fast_spectral_methods__deriving_numerical_parameters}
\begin{split}
    &\hat{G}(l,m)\\
    &=
    \int_0^{2S} \int_{S^{d-1}} \int_{S^{d-1}}
    B(r, \sigma \cdot \omega)
    e^{ -i \frac{\pi}{2T} r \omega \cdot \frac{1+e}{2}(l+m) }
    e^{  i \frac{\pi}{2T} r \sigma \cdot \frac{1+e}{2}l }
    e^{ -i \frac{\pi}{2T} r \sigma \cdot \frac{3-e}{2}m }
    \, d\omega \, d\sigma \, dr,\\
    &=
    \int_0^{2S} \int_{S^{d-1}} {
        F(l+m; r, \sigma)
        e^{  i \frac{\pi}{2T} r \sigma \cdot \frac{1+e}{2}l }
        e^{ -i \frac{\pi}{2T} r \sigma \cdot \frac{3-e}{2}m }
    } \, d\sigma \, dr,
\end{split}
\end{align}
where
\begin{align}
    F(k; r, \sigma)
    \coloneqq
    \int_{S^{d-1}} {
        B(r, \sigma \cdot \omega) \exp\left( -i \frac{\pi}{2T} r \omega \cdot \frac{1+e}{2}k \right)
    } \, d\omega
\end{align}
for all $k \in \mathbb{Z}^d$ and $\sigma, \omega \in S^{d-1}$.
\reviewerX{Let $\{ r_a \}_{a = 1}^{N_r},\, \{ \sigma_b \}_{n=1}^{N_\sigma}$ be the sets of quadrature points for the integration with respect to $r$ and $\sigma$, and $\{ \Delta{r_a} \}_{a=1}^{N_a}$ and $\{ \Delta{\sigma_b} \}_{b=1}^{N_\sigma}$ be the sets of quadrature points for the integration with respect to $r$ and $\sigma$, respectively, with $N_r N_\sigma \ll N^{d-1}$.}
Then $\hat{G}(l, m)$ can be approximated as
\begin{align*}
\begin{split}
    \hat{G}(l, m)
    \approx
    \hat{G}^{\text{fast}}(l, m)
    \coloneqq
    \sum_{\substack{1 \leq a \leq N_r\\1 \leq b \leq N_\sigma}} {
        \gamma_{a, b}(l+m) \alpha_{a, b}(l) \beta_{a, b}(m)
    },
\end{split}
\end{align*}
where $\alpha_{a, b}(l)\coloneqq\exp\left(  i \frac{\pi}{2T} r_a \sigma_b \cdot \frac{1+e}{2} l \right)$, $\beta_{a, b}(m)\coloneqq\exp\left( -i \frac{\pi}{2T} r_a \sigma_b \cdot \frac{3-e}{2} m \right)$, and $\gamma_{a, b}(l+m)\coloneqq F(l+m; r_a, \sigma_b) \Delta r_a \Delta \omega_b$.
% \begin{align*}
%     \alpha_{a, b}(l)
%     &\coloneqq
%     \exp\left(  i \frac{\pi}{2T} r_a \sigma_b \cdot \frac{1+e}{2} l \right),
%     \\
%     \beta_{a, b}(m)
%     &\coloneqq
%     \exp\left( -i \frac{\pi}{2T} r_a \sigma_b \cdot \frac{3-e}{2} m \right),
%     \\
%     \gamma_{a, b}(l+m)
%     &\coloneqq
%     F(l+m; r_a, \sigma_b) \Delta r_a \Delta \omega_b.
% \end{align*}
Up to rearrangement, $\hat{G}^{\textrm{fast}}(l, m)$ can be written as
\reviewerX{
    \begin{equation}\label{eq:decoupled_G}
        \hat{G}^{\text{fast}}(l,m)
        =
        \sum \limits_{t=1}^{M_\text{quad}-1} \gamma_t(l+m) \alpha_t(l) \beta_t(m),
    \end{equation}
}
with \reviewerX{$M_\text{quad}-1 = N_r N_\sigma$}.
Defining \reviewerX{$\alpha_{M_\text{quad}}(m) = 1$, $\beta_{M_\text{quad}}(m) = -\hat{G}^\text{fast}(m, m)$, $\gamma_{M_\text{quad}}(m)=1$} for all $m \in \mathbb{Z}^d$ with $\| m \|_\infty \leq N$ and \reviewerX{$\alpha_{M_\text{quad}}(m) = \beta_{M_\text{quad}}(m) = \gamma_{M_\text{quad}}(m) = 0$ otherwise}, it follows that
\begin{equation}\label{eq:fast_spectral_method__loss_integrated_into_gain}
    \begin{split}
        %%%%%
        \hat{Q}_k^{\text{fast}}(f_N,f_N)
        &=\sum \limits_{\substack{\| l \|_\infty, \| m \|_\infty \leq N \\ l+m=k}}^{} \left( \hat{G}^\text{fast}(l, m) - \hat{G}^\text{fast}(m, m) \right) \hat{f}_l \hat{f}_m
        \\
        %%%%%
        &=
        \sum \limits_{\substack{\| l \|_\infty, \| m \|_\infty \leq N \\ l+m=k}}^{} \left( \sum \limits_{t=1}^{\reviewerX{M_\text{quad}}} \gamma_t(l+m) \alpha_t(l) \beta_t(m) \right) \hat{f}_l \hat{f}_m
        \\
        %%%%%
        &=
        \sum \limits_{t=1}^{\reviewerX{M_\text{quad}}} \gamma_t(k) \sum \limits_{\substack{\| l \|_\infty, \| m \|_\infty \leq N \\ l+m=k}}^{} \left( \alpha_t(l) \hat{f}_l \right) \left( \beta_t(m) \hat{f}_m \right).
    \end{split}
\end{equation}
Therefore, $\hat{Q}_k$ can be efficiently approximated from $\hat{f}_k$ using the decomposition of the form
\begin{align}\label{eq:fastfourier_decompose}
\hat{Q}_k^{\text{fast}}(f_N,f_N) = \sum_{t=1}^{\reviewerX{M_\text{quad}}} \left[ \gamma_t(k) \cdot \left( (\alpha_t \hat{f}) * (\beta_t \hat{f}) \right)_k \right],
\end{align}
where $\cdot$ denotes the elementwise multiplication and $*$ denotes the discrete convolution.
Note that the discrete convolutions can be computed using the FFT. Specifically, each convolution term is evaluated via
\begin{equation}\label{eqn:internal_convolution_using_fft}
\begin{split}
    (\alpha_t \hat{f}) * (\beta_t \hat{f}) = \mathrm{FFT} \left[ \mathrm{FFT}^{-1}(\alpha_t \hat{f}) \cdot \mathrm{FFT}^{-1}(\beta_t \hat{f}) \right],
\end{split}
\end{equation}
and for any function $u: \mathbb{Z}^d \rightarrow \mathbb{C}$, $\mathrm{FFT}^{-1}(u)$ is computed using the values $u(k)$ for $k \in \mathbb{Z}^d$ satisfying $\| k \|_\infty \leq N$.
This FFT-based convolution allows for an efficient approximation of $Q$ while preserving spectral accuracy.
\reviewerY{
    To be specific, the convolution in (\ref{eqn:internal_convolution_using_fft}) can be computed in $O(N^d \log{N})$ operations for each $t$, so the fast spectral method requires $O(M N^d \log{N})$ operations for computing the collision term.
    However, as the integrand in (\ref{eq:fast_spectral_methods__deriving_numerical_parameters}) osciallates on the scale of $\norm{\frac{1+e}{2}l-\frac{3-e}{2}m}_{2}$, the number $N_r$ of the quadrature points must be at least $O(N)$, as explained in \cite{MR3687853}.
    Therefore, the fast spectral method for solving the Boltzmann equation requires $O(M N^{d+1} \log{N})$ operations, where $M = N_\sigma$.
}
Building upon this efficient spectral decomposition, we now introduce our proposed model, the Fourier Neural Spectral Network, which integrates the Fourier spectral method with networks to learn the collision operator in a resolution-invariant manner.

\section{Proposed model: Fourier neural spectral network (FourierSpecNet)}\label{sec:method} 
We propose a novel framework for solving the Boltzmann equation, called the Fourier Neural Spectral Network (FourierSpecNet), which combines the advantages of the Fourier spectral method rooted in numerical analysis with deep learning. This approach aims to leverage the strengths of both methodologies. \reviewerY{Unlike conventional neural architectures that are designed heuristically, FourierSpecNet is directly derived from the analytical structure of the fast Fourier spectral method for the Boltzmann collision operator. In the classical fast spectral method, the operator is represented through separable convolutional terms $\{\alpha_t, \beta_t, \gamma_t\}_{t=1}^{M_{\rm{quad}}}$ that determine weighted convolutions in the Fourier domain. 
The FourierSpecNet inherits this mathematical structure but replaces the deterministic coefficients with their trainable neural counterparts $\{\alpha^{nn}_t, \beta^{nn}_t, \gamma^{nn}_t\}_{t=1}^{M}$, which are learned from data while preserving the separable form. Hence, the proposed model is not a new deep architecture in the conventional sense, but rather a numerically structured operator network whose form is motivated by the spectral decomposition of the Boltzmann collision operator. 
This design allows FourierSpecNet to maintain the spectral accuracy and stability of the classical solver, while enabling data-driven learning of the spectral weights within a rigorously defined mathematical framework.}

\begin{figure}[]
    \includegraphics[width=\textwidth]{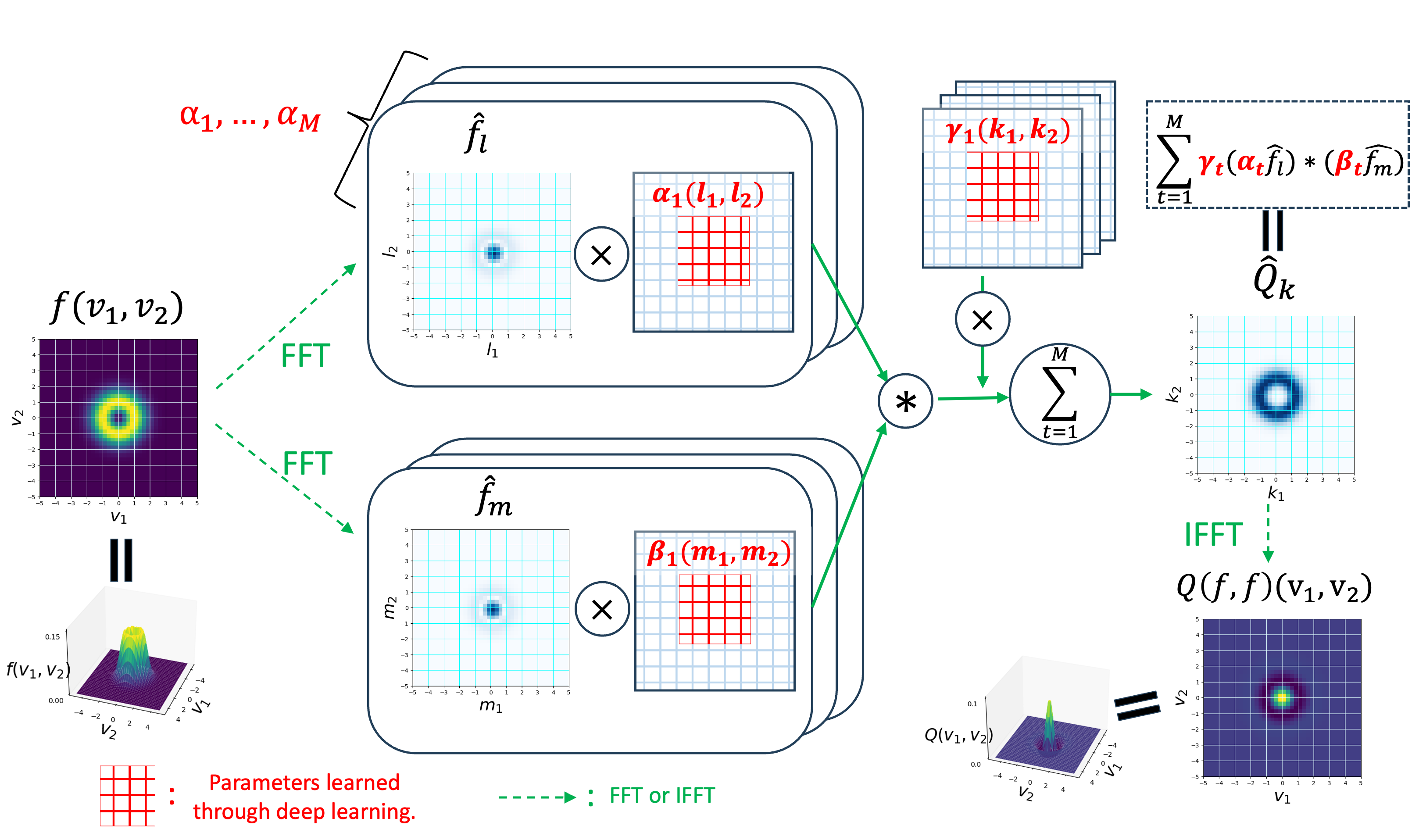}
    \caption{Overview of the proposed Fourier neural spectral network to approximate the collision operator of the Boltzmann equation. Note that the size of learnable network parameters $\{\alpha^{nn}_t\}_{t=1}^M$, $\{\beta^{nn}_t\}_{t=1}^M$, and $\{\gamma^{nn}_t\}_{t=1}^M$ are independent of the size of input $f(\boldsymbol{v})$.}
    \label{fig:framework}
\end{figure}

The diagram of the proposed FourierSpecNet is explained in Figure \ref{fig:framework}.
The key idea of FourierSpecNet is to learn the weight $\hat{G}(l,m)$ in \eqref{eq:fourier_weight} using deep learning.
Following the Fourier spectral method described in the preceding section, we discretize $f(\boldsymbol{v})$ and perform FFT to obtain $\hat{f}_l$ and $\hat{f}_m$. Utilizing the decoupled form \eqref{eq:decoupled_G} proposed in~\cite{MR3687853}, we take the complex-valued neural network parameters $\{\alpha^{nn}_t(k), \beta^{nn}_t(k), \gamma^{nn}_t(k): k \in \mathbb{Z}^d, t=1,\cdots,M\} \subset \mathbb{C}$. Then, we approximate $\hat{Q}_k$ in a manner similar to \eqref{eq:fastfourier_decompose} as follows:
\begin{align}\label{eq:nn_decompose}
\begin{split}
    \hat{Q}^{nn}_k
    \coloneqq
    \sum_{t=1}^{M} \left[
        \gamma^{nn}_t(k) \left(
            (\alpha^{nn}_t \hat{f})
            *
            (\beta^{nn}_t \hat{f})
        \right)_k
    \right].
\end{split}
\end{align}

Finally, applying the inverse Fourier transform to $\hat{Q}^{nn}_k$ yields the approximated collision operator $Q^{nn}$, representing the output of the proposed FourierSpecNet to approximate the collision operator $Q$.
Importantly, the neural network parameters $\alpha^{nn}_t$, $\beta^{nn}_t$, and $\gamma^{nn}_t$ are constrained to the pre-defined $N_\textrm{trun}^d$ lowest frequency modes, where $N_{\text{trun}}(\leq N)$ is a positive integer independent of $N$.
To be precise, we set $\alpha^{nn}_t(k) = \beta^{nn}_t(k) = \gamma^{nn}_t(k) = 0$ whenever $k \in \mathbb{Z}^d$ and $\| k \|_{\infty} > N_\mathrm{trun}$.
This provides a significant advantage, as once FourierSpecNet is trained, it allows for the prediction of $Q(f, f)$ irrespective of the discretization of the input $f(\boldsymbol{v})$. The details are in Section \ref{subsec:super_resol}.

\reviewerY{
Each neural component $\alpha^{nn}_t$, $\beta^{nn}_t$, and $\gamma^{nn}_t$ can be viewed as a learnable spectral coefficients that modulates the Fourier modes of $\hat{f}$ in a separable manner. These are trained end-to-end using paired data $(f, Q(f,f))$, and their role corresponds to learning the underlying kernel $\widehat{G}(l,m)$ that defines the Boltzmann collision operator in spectral space. This formulation preserves the mathematical interpretability of the fast spectral method while introducing flexibility to capture complex, data-driven corrections to the analytical kernel.
}

\reviewerY{
    % Regarding the zero-shot superresolution computational time
    % A notable property of FourierSpecNet compared to the fast spectral method is that the number of required parameters $\{\alpha^{nn}_t, \beta^{nn}_t, \gamma^{nn}_t\}_{t=1}^{M}$ is independent of the input resolution $N$.
    A notable advantage of FourierSpecNet over the fast spectral method lies in its resolution-independent parameterization.
    The parameters $\{\alpha^{nn}_t, \beta^{nn}_t, \gamma^{nn}_t\}_{t=1}^{M}$ are defined only on truncated Fourier space, and their number is determined as a hyperparameter rather than being dependent on the input resolution $N$ as in the fast spectral method.
    Consequently, once trained on lower-resolution distributions, the same parameters can be directly applied to infer the solutions for higher-resolution inputs without retraining.
    This design yields a total computational cost of $O(M N^d \log{N})$, which is more efficient than the $O(M N^{d+1} \log{N})$ complexity of the fast spectral method, where the separable rank grows with $N$ due to the quadrature-based summation.
}

\subsection{Construction of Training Dataset for FourierSpecNet}
First, we need the training dataset to train the FourierSpecNet. Determining the training data is a crucial aspect of training neural network models, and the selection of training data for each task significantly influences the model's generalization and performance. To solve the Boltzmann equation using FourierSpecNet, we generate training data that aligns with this objective. Specifically, drawing on the asymptotic convergence of solutions of the well-known Boltzmann equation to its Maxwellian, as established in existing PDE theories, we create input functions using Maxwellians and their perturbations. Utilizing this approach similar to~\cite{MR4554720,MR4609518}, we construct appropriate training input data $\mathcal{D} \coloneqq \{M_i\}_{i=1}^{1000}\cup\{M^2_i\}_{i=1}^{1000}\cup\{M^p_i\}_{i=1}^{1000}$ to effectively learn the relationship between solutions $f(\boldsymbol{v})$ evolving over time in the Boltzmann equation and their corresponding collision operators $Q(f, f)$ as follows:
\begin{itemize}
     \item (Gaussian function) $\{M_i\}_{i=1}^{1000}$ where
     \begin{equation}\notag
         M_i(\boldsymbol{v})\eqdef\frac{1}{(2\pi\sigma_i^2)^{d/2}}\exp\left(-\frac{|\boldsymbol{v}-\boldsymbol{c}_i|^2}{2\sigma_i^2}\right)
     \end{equation}with $\boldsymbol{c}_i\in[-1,1]^d$ and $\sigma_i\in[0.8,1.2]$.
     \item (Sum of two Gaussian functions) $\{M^2_i\}_{i=1}^{1000}$ where
     \begin{equation}\notag
         M^2_i(\boldsymbol{v})\eqdef M_{i_1}(\boldsymbol{v})+M_{i_2}(\boldsymbol{v}),
     \end{equation}where $M_{i_1}$ and $M_{i_2}$ are also random Gaussian functions.
     \item (Gaussian function with a small perturbation) $\{M^p_i\}_{i=1}^{1000}$ where
     \begin{equation}\notag
         M^p_i(\boldsymbol{v})\eqdef M_i(\boldsymbol{v})+M_i(\boldsymbol{v})g(\boldsymbol{v}),
     \end{equation}where $g$ is a polynomial of degree 2 with randomly chosen coefficients in $[0,1]$. 
\end{itemize}
Each input function $f\in\mathcal{D}$ is normalized so that the volume of the function is 1 for training stability.
% Furthermore, we use the loss function.
The target collision operator $Q(f,f)$ for each $f\in\mathcal{D}$ is obtained by the Fourier spectral method~\cite{MR3687853}, and the FourierSpecNet is trained via supervised learning using the following relative $L^2$ loss function:
\begin{equation}
    \text{Loss}=\frac{1}{|\mathcal{D}|}\sum_{f\in\mathcal{D}}\frac{ \| Q^{nn}(f,f)-Q(f,f) \|_{L^2(D_T)}}{\|Q(f,f) \|_{L^2(D_T)}}.
\end{equation}

\subsection{FourierSpecNet with zero-shot super-resolution capabilities}\label{subsec:super_resol}
\begin{figure}[]
\centering
  \includegraphics[width=0.7\textwidth]{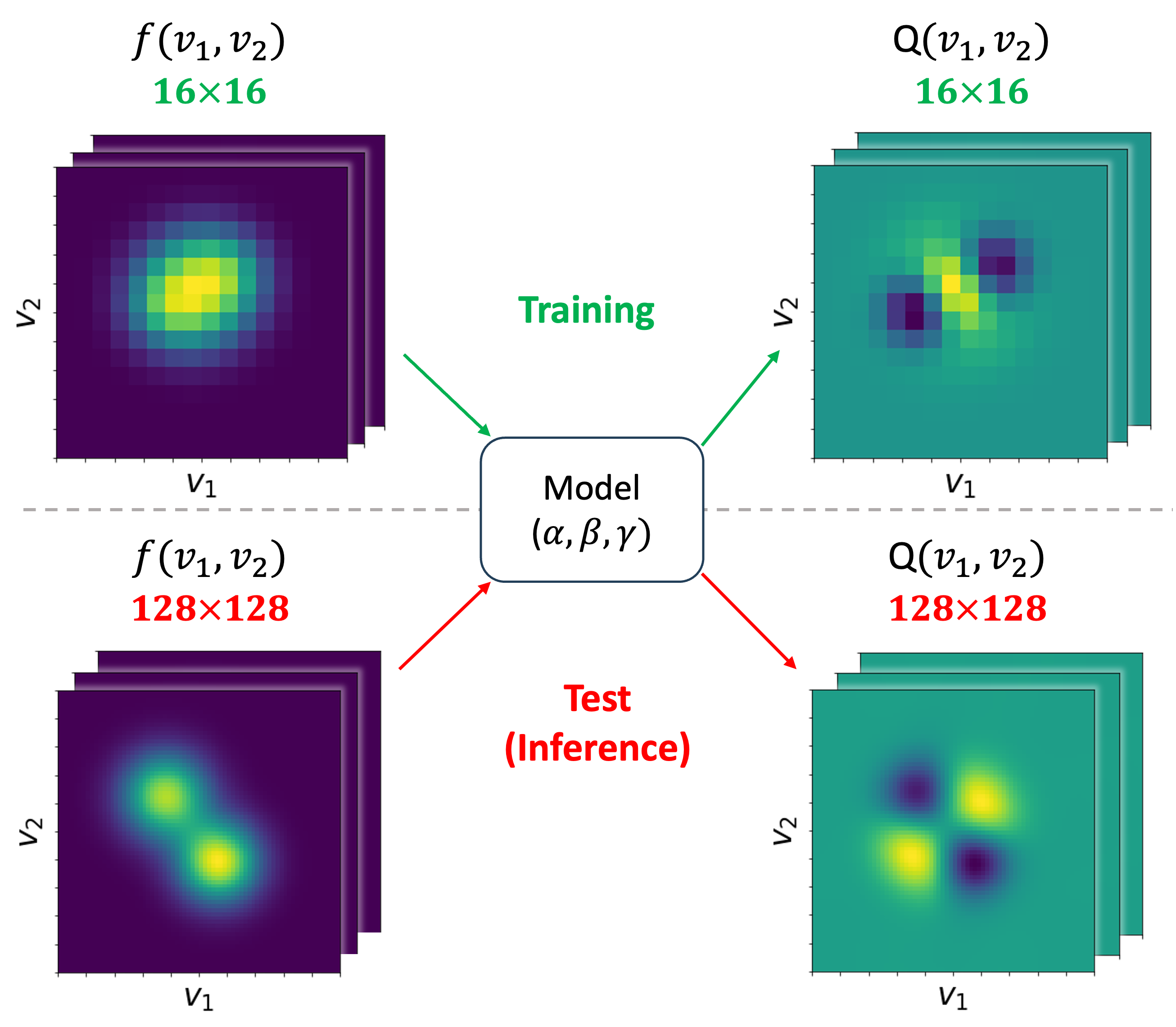}
  \caption{Illustration of the proposed Fourier Neural Spectral Network as a super-resolution approximator for the Boltzmann collision operator. The model ($\alpha, \beta, \gamma$) is trained on low-resolution input-output pairs ($f(v_1, v_2)$ and $Q(v_1, v_2)$) defined on a $16 \times 16$ grid. During inference, the model generalizes to high-resolution inputs ($128 \times 128$ grid) and accurately predicts the corresponding high-resolution outputs, demonstrating its resolution-invariance and super-resolution capabilities.}
  \label{fig:superresol}
\end{figure}

One of the most significant advantages of FourierSpecNet is its super-resolution capability. Similar to~\cite{li2021fourier}, the neural network parameters $\alpha^{nn}_t$, $\beta^{nn}_t$, and $\gamma^{nn}_t$ have a size of $N_{\text{trun}}^d$, which is independent of the discretization size $N^d$ for the input function $f(\boldsymbol{v})$. This implies that FourierSpecNet can use the same neural network parameters regardless of the resolution of the input function $f(\boldsymbol{v})$. \reviewerX{In contrast to the classical fast spectral method, where the number of separable terms (often denoted by $M_{\mathrm{quad}}$) scales with the number of quadrature points used in the angular integration, the parameter count in FourierSpecNet is fixed by the truncated spectral bandwidth $N_{\mathrm{trun}}$ and does not depend on the discretization grid $N^d$. Here, $M$ in our model represents the number of learned separable spectral modes, which serves as a neural hyperparameter and should not be confused with $M_{\mathrm{quad}}$ from the quadrature-based formulation.} It is noteworthy that traditional Fourier spectral methods and their variants are constrained by the discretization size of the velocity domain, limiting the obtainable $Q(\boldsymbol{v})$ to the same resolution as the input function $f(\boldsymbol{v})$. In contrast, FourierSpecNet is a resolution-invariant model, enabling the evaluation of $Q(f,f)$ at a higher resolution after the model is trained on a lower resolution. The super-resolution advantage of the proposed FourierSpecNet is illustrated in Figure \ref{fig:superresol}.

\subsection{Properties of the Fourier neural spectral network}
In this section, we demonstrate the properties of the proposed method. The Fourier neural spectral network leverages the spectral method numerical analysis technique by integrating it with deep learning. Consequently, it inherits significant strengths from the well-established properties within traditional spectral methods to solve the Boltzmann equation. For example, the approximated physical quantities using $f_N$ from \eqref{eq:trunk_f} defined as
\begin{equation}
    \rho_N:=\int_{\mathbb{R}^d} f_N d\boldsymbol{v}, \; \boldsymbol{u}_N:=\frac{1}{\rho_N}\int_{\mathbb{R}^d} f_N\boldsymbol{v} d\boldsymbol{v}, \; T_N:=\frac{1}{d\rho_N}\int_{\mathbb{R}^d} f_N|\boldsymbol{u}-\boldsymbol{v}|^2 d\boldsymbol{v}.
\end{equation}
has the following properties:
\begin{itemize}
    \item{
        $\rho = \rho_N$ always holds whatever $N$ is.
    }
    \item{
        There exist constant $C_1, C_2$ which depend only on $T$ and satisfy
        \begin{equation}
            |M - M_N|
            \leq
            \frac{C_1}{N^{1/2}} \| f \|_{L^2(D_T)},
            \;
            |E - E_N|
            \leq \frac{C_2}{N^{3/2}} \| f \|_{L^2(D_T)}.
        \end{equation}
    }
\end{itemize}
More properties and details are explained in Section 5 at~\cite{MR3202241}.
Besides the properties of physical quantities, FourierSpecNet also admits a theoretical consistency guarantee. Let $Q_N^{\text{fast}}$ denote an approximation of the truncated collision operator $Q_N$ using $\hat{Q}_k^{\text{fast}}$ in \eqref{eq:fastfourier_decompose}.
The following proposition provides an upper bound on the approximation error of the trained collision operator in FourierSpecNet.
\begin{proposition}[Consistency of FourierSpecNet]\label{prop:consistency_gjjung}
Suppose $r \geq 0$ and the collision kernel satisfies $\hat{G}(l, m) \rightarrow 0$ as $\| l \|_\infty \rightarrow \infty$ or $\| m \|_\infty \rightarrow \infty$. Then for any $\varepsilon > 0$, there exist
\begin{enumerate}
\item[(\romannumeral 1)] positive integers $M, N_{\mathrm{trun}} \in \mathbb{N}$,
\item[(\romannumeral 2)] coefficients $\{\alpha^{nn}_t\}_{t=1}^{M}, \{\beta^{nn}_t\}_{t=1}^{M}, \{\gamma^{nn}_t\}_{t=1}^{M} \subset \mathbb{C}^{N_\mathrm{trun}^d}$,
\end{enumerate}
such that for any $f \in L^\infty(D_T)$ and any $N \in \mathbb{N}$, the following error bound holds:
\begin{align}\label{eq:consistency__gjjung}
    \begin{split}
        &\|Q(f, f) - Q^{nn}(f_N, f_N) \|_{L^2(D_T)}
        \\
        &\leq
        C\left(
            \| f - f_N \|_{L^2(D_T)}
            +
            \frac{1}{N^r} \| Q(f_N, f_N) \|_{H^r(D_T)}
        \right)
        +
        \varepsilon \| f \|_{L^\infty(D_T)}^2,
    \end{split}
\end{align}
where $C > 0$ is a constant depending on $T$ and $\| f \|_{L^2(D_T)}$.
Furthermore, if $f \in L^\infty(D_T) \cap H^r(D_T)$, then
\begin{align}\label{eq:consistency__gjjung__2}
    \begin{split}
        &\|Q(f, f) - Q^{nn}(f_N, f_N) \|_{L^2(D_T)}
        \\
        &\leq
        \frac{C'}{N^r}\left(
            \| f \|_{H^r(D_T)}
            +
            \| Q(f_N, f_N) \|_{H^r(D_T)}
        \right)
        +
        \varepsilon \| f \|_{L^\infty(D_T)}^2
    \end{split}
\end{align}
for some constant $C'>0$ depending on $T$ and $\| f \|_{L^2(D_T)}$.
\end{proposition}
\begin{remark*}
In Appendix~\ref{sec:decay_of_kernel_modes}, it is proved that $\hat{G}(l, m) \rightarrow 0$ as $\| l \|_\infty \rightarrow \infty$ or $\| m \|_\infty \rightarrow \infty$ for the variable hard-sphere (VHS) model when the inelasticity coefficient $e$ takes a rational value.
\end{remark*}

To establish Proposition~\ref{prop:consistency_gjjung}, we utilize the following sequence of lemmas, which provide key intermediate bounds related to the spectral approximation and neural network-based parameterization.

%%%%% Error from the approximation of the spectral method
\begin{lemma}\label{lemma:consistency__gjjung_1}
    Assume that $\hat{G}(l, m) \rightarrow 0$ as $\| l \|_\infty + \| m \|_\infty \rightarrow \infty$.
    For each $\varepsilon > 0$, there are $M \in \mathbb{N}$ and compactly supported functions $\alpha_t,\, \beta_t,\, \gamma_t: \mathbb{Z}^d \rightarrow \mathbb{C}$ for $t = 1, \cdots, M$ satisfying
    \begin{align*}
        \| Q_N(f_N,f_N) - Q_N^{\mathrm{fast}}(f_N,f_N) \|_{L^2(D_T)}
        \leq
        \varepsilon \| f \|_{L^\infty(D_T)}^2
    \end{align*}
    for all $f \in L^\infty(D_T)$ and $N \in \mathbb{N}$.
\end{lemma}
\begin{proof}
    Recall from \eqref{eq:fast_spectral_method__loss_integrated_into_gain} that
    
    $\displaystyle{
        \hat{Q}_k^{\text{fast}}(f_N,f_N)
        =
        \sum_{\substack{l, m \in \mathbb{Z}^d \\ l+m=k}}{
            (\hat{G}^{\text{fast}}(l,m) + \hat{g}^{\text{fast}}(m)) \hat{f}_l \hat{f}_m
        }
    }$ 
    for all $k \in \mathbb{Z}^d$ with $\| k \|_\infty \leq N$, where $\hat{g}^\text{fast}(m) \coloneqq -\hat{G}^\text{fast}(m,m)$ for all $m \in \mathbb{Z}^d$.
    With the $l^p$ norm on $l^p(\mathbb{Z}^d)$ defined as $\| x \|_{l^p(\mathbb{Z}^d)} = \left( \sum_{j \in \mathbb{Z}^d} |x_j|^p \right)^{1/p}$ for all $p \in [1, \infty)$ and $x \in l^p(\mathbb{Z}^d)$, Parseval's identity implies
    \begin{align*}
        &\| Q_N(f_N,f_N)-Q_N^{\text{fast}}(f_N,f_N) \|_{L^2(D_T)}^2
        =
        (2T)^d
        \| \hat{Q}_N(f_N,f_N) - \hat{Q}_N^{\text{fast}}(f_N,f_N) \|_{l^2(\mathbb{Z}^d)}^2
        \\
        &=
        (2T)^d
        \sum_{\substack{k \in \mathbb{Z}^d, \\ |k| \leq N}} \bigg|{
            \sum_{\substack{l, m \in \mathbb{Z}^d, \\ l+m=k}} {
                \left(
                    \left( \hat{G}(l, m) - \hat{G}^{\text{fast}}(l, m) \right)
                    +
                    \left( \hat{g}(m) - \hat{g}^\text{fast}(m) \right)
                \right)
                \hat{f}_l \hat{f}_m
            }
        }\bigg|^2
        \\
        &\leq
        (2T)^d
        \bigg[
            \sum_{k \in \mathbb{Z}^d} {
                \sum_{\substack{l, m \in \mathbb{Z}^d \\ l+m=k}} {
                    \left(
                        \left| \hat{G}(l, m) - \hat{G}^{\text{fast}}(l, m) \right|
                        +
                        \left| \hat{g}(m) - \hat{g}^{\text{fast}}(m) \right|
                    \right)
                    \left| \hat{f}_l \hat{f}_m \right|
                }
            }
        \bigg]^2.
    \end{align*}
    Therefore, we have
    \begin{multline}
        \| Q_N(f_N,f_N) - Q_N^{\text{fast}}(f_N,f_N) \|_{L^2(D_T)}\\
        \lesssim
        \sum_{l, m \in \mathbb{Z}^d} {
            \left(
                \left| \hat{G}(l, m) - \hat{G}^{\text{fast}}(l, m) \right|
                +
                \left| \hat{g}(m) - \hat{g}^{\text{fast}}(m) \right|
            \right)
            \left| \hat{f}_l \hat{f}_m \right|
        }.
    \end{multline}
    Recall that
    $
        \hat{G}^{\text{fast}}(l, m)
        =
        \sum \limits_{t=1}^{M-1} {
            \gamma_t(l+m) \alpha_t(l) \beta_t(m)
        }
    $
    for all $l, m \in \mathbb{Z}^d$.
    Since $\hat{G}(l, m) \rightarrow 0$ as $\| l \|_\infty \rightarrow \infty$ or $\| m \|_\infty \rightarrow \infty$, there is a positive integer $K$ for which $| \hat{G}(l, m) | < \varepsilon/2$ for all $l, m \in \mathbb{Z}^d$ satisfying $\| l \|_\infty > K$ or $\| m \|_\infty > K$.
    For the case $l, m \in \mathbb{Z}^d$ with $\| l \|_\infty,\, \| m \|_\infty \leq K$, by using numerical integration with a sufficiently fine grid in the estimation of \eqref{eq:fast_spectral_methods__deriving_numerical_parameters}, one can find $M \in \mathbb{N}$ and $\alpha_t(l),\, \beta_t(m),\, \gamma_t(l+m) \in \mathbb{C}$ for $t=1, \cdots, M-1$, for which $\left| \hat{G}(l, m) - \hat{G}^{\text{fast}}(l, m) \right| < \varepsilon/2$.
    Next, if $\alpha_t(j)$ and $\beta_t(j)$ are set 0 for $t=1, \cdots, M-1$ and $j \in \mathbb{Z}^d$ satisfying $\| j \|_\infty > K$, we have $\hat{G}^{\text{fast}}(l, m) = 0$ and $\left| \hat{G}(l, m) - \hat{G}^{\text{fast}}(l, m) \right| < \varepsilon/2$ for all $l, m \in \mathbb{Z}^d$ with $\| l \|_\infty > K$ or $\| m \|_\infty > K$. Moreover, if $\gamma_t(j) = 0$ for $t=1, \cdots, M-1$ and $j \in \mathbb{Z}^d$ with $\| j \|_\infty > 2K$, then $\alpha_t,\, \beta_t,\, \gamma_t\, (t = 1, \cdots, M-1)$ are compactly supported. We then assign the values of $\alpha_t(l),\, \beta_t(m),\, \gamma_t(l+m)$ which have not been determined arbitrarily for $t=1, \cdots, M-1$.
    Finally, we define $\alpha_M(m) = 1$, $\beta_M(m) = -\hat{G}^\text{fast}(m, m)$, $\gamma_M(m) = 1$ for $m \in \mathbb{Z}^d$ with $\| m \|_\infty \leq K$, and $\alpha_M(m) = \beta_M(m) = \gamma_M(m) = 0$ otherwise.
    Then $\alpha_M, \beta_M, \gamma_M$ are also compactly supported, and $\left| \hat{g}(m) - \hat{g}^\text{fast}(m) \right| < \epsilon/2$ for all $m \in \mathbb{Z}^d$.
    Then, by the Hausdorff-Young inequality,
    \begin{multline}
        \| Q_N(f_N,f_N) - Q_N^{\text{fast}}(f_N,f_N) \|_{L^2(D_T)}\\
        \lesssim
        \left( \frac{\varepsilon}{2}+\frac{\varepsilon}{2} \right) \sum_{l, m \in \mathbb{Z}^d} {
            \left|\hat{f}_l \hat{f}_m\right|
        }
        =
        \varepsilon \| \hat{f} \|_{l^1(\mathbb{Z}^d)}^2
        \lesssim
        \varepsilon \| f \|_{L^\infty(D_T)}^2.
    \end{multline}
    This completes the proof.
\end{proof}
%%%%%%%%%%%%%%%%%%%%%%%%%%%%%%%%%%%%%%%%%%%%%%%%%%

%%%%% Error from the approximation of the kernel modes in the fast spectral method
\begin{lemma}\label{lemma:consistency__gjjung_2}
    Let $M \in \mathbb{N}$ be given,
    \reviewerX{
        and define
        \begin{align*}
            A_t
            &\coloneqq
            \norm{\gamma_t}_{l^4(\mathbb{Z}^d)} \left(1 + \norm{\alpha_t}_{l^2(\mathbb{Z}^d)} + \norm{\beta_t}_{l^2(\mathbb{Z}^d)}\right),
            \\
            B_t
            &\coloneqq
            \left( 1 + \norm{\alpha_t}_{l^2(\mathbb{Z}^d)} \right)
            \left( 1 + \norm{\beta_t}_{l^2(\mathbb{Z}^d)} \right)
            \left( 1 + \norm{\gamma_t}_{l^4(\mathbb{Z}^d)} \right)
        \end{align*}
        for $t = 1, \cdots, M$, and $\displaystyle{c \coloneqq \max_{1\leq t \leq M}(A_t+B_t)}$}.
        For each $\varepsilon > 0$, there is $\delta > 0$ \reviewerX{with $\displaystyle{\delta = O\left( \varepsilon / (cM) \right)}$} for which
        \begin{center}
            $\|  Q_N^{\rm{fast}}(f_N,f_N) - Q^{nn}(f_N,f_N) \|_{L^2(D_T)} \leq \varepsilon \| f \|_{L^\infty(D_T)}^2$
            for all \reviewerX{$f \in L^\infty(D_T)$},
        \end{center}
        whenever
        \begin{align*}
            \| \alpha_t - \alpha^{nn}_t \|_{l^2(\mathbb{Z}^d)} < \delta,
            \,
            \| \beta_t - \beta^{nn}_t \|_{l^2(\mathbb{Z}^d)} < \delta,
            \,
            \| \gamma_t - \gamma^{nn}_t \|_{l^4(\mathbb{Z}^d)} < \delta.
        \end{align*}
\end{lemma}
\reviewerX{\begin{remark*}
    Lemma~\ref{lemma:consistency__gjjung_2} explains how close the parameters of the FourierSpecNet must be to the parameters in the fast spectral method in order for the FourierSpecNet to satisfy
    \begin{align*}
        \norm{
            Q^{\mathrm{fast}}_N(f_N, f_N) - Q^{nn}(f_N, f_N)
        }_{L^2(D_T)}
        \leq
        \varepsilon \norm{f}_{L^{\infty}(D_T)}^2
    \end{align*}
    for all $f \in L^\reviewerX{\infty}(D_T)$.
    Specifically, Lemma~\ref{lemma:consistency__gjjung_2} states that
    \begin{align*}
        \| \alpha_t - \alpha^{nn}_t \|_{l^2(\mathbb{Z}^d)},\,
        \| \beta_t - \beta^{nn}_t \|_{l^2(\mathbb{Z}^d)},\,
        \| \gamma_t - \gamma^{nn}_t \|_{l^4(\mathbb{Z}^d)}
    \end{align*}
    should be less than $\delta>0$ for $t = 1, \cdots, M$, where $\delta$ scales with $\varepsilon$ and scales inversely with $M$.
\end{remark*}}
\begin{proof}
    Assume \reviewerX{$f \in L^\infty(D_T)$}.
    Since
    \begin{align*}
        &\left|\left( \hat{Q}^\text{fast}_N(f_N, f_N) - \hat{Q}^{nn}(f_N, f_N) \right)_k\right|^2
        \\
        &=
        \left|
            \sum_{t=1}^{M} \left[
                \gamma_t(k)
                \left\{
                    (\alpha_t \hat{f}) * (\beta_t \hat{f})
                \right\}_k
                -
                \gamma^{nn}_t(k)
                \left\{
                    (\alpha^{nn}_t \hat{f}) * (\beta^{nn}_t \hat{f})
                \right\}_k
            \right]
        \right|^2
        \\
        &\leq
        M
        \sum_{t=1}^{M}
        \left|
            \left[
                \gamma_t(k)
                \left\{
                    (\alpha_t \hat{f}) * (\beta_t \hat{f})
                \right\}_k
                -
                \gamma^{nn}_t(k)
                \left\{
                    (\alpha^{nn}_t \hat{f}) * (\beta^{nn}_t \hat{f})
                \right\}_k
            \right]
        \right|^2
    \end{align*}
    for $k \in \mathbb{Z}^d$, by Parseval's identity, we have
    \begin{align*}
        &\| Q_N^{\text{fast}}(f_N,f_N) - Q^{nn}(f_N,f_N) \|_{L^2(D_T)}^2
        =
        (2T)^d
        \| \hat{Q}_N^{\text{fast}}(f_N,f_N) - \hat{Q}^{nn}(f_N,f_N) \|_{l^2(\mathbb{Z}^d)}^2
        \\
        &\leq
        M (2T)^d
        \sum_{t=1}^{M} \sum_{k\in\mathbb{Z}^d} \left|
            \gamma_t(k) \left\{ ((\alpha_t \hat{f}) * (\beta_t \hat{f}) \right\}_k
            -
            \gamma^{nn}_t(k) \left\{ (\alpha^{nn}_t \hat{f}) * (\beta^{nn}_t \hat{f}) \right\}_k
        \right|^2,
    \end{align*}
    whence
    \reviewerX{\begin{align}
        \label{eq:consistency__gjjung_3__ineq_main_1}
        \begin{split}
            \|
                Q_N^{\text{fast}}(f_N,f_N)
                -
                Q^{nn}(f_N,f_N)
            \|_{L^2(D_T)}^2
            \leq
            M(2T)^d
            \sum_{t=1}^{M} \|
                \gamma_t \zeta_t
                -
                \gamma^{nn}_t \zeta^{nn}_t
            \|_{l^2(\mathbb{Z}^d)}^2,
        \end{split}
        \end{align}
    }
    where
    $\zeta_t \coloneqq (\alpha_t \hat{f}) * (\beta_t \hat{f})$
    and
    $\zeta^{nn}_t \coloneqq (\alpha^{nn}_t \hat{f}) * (\beta^{nn}_t \hat{f})$.
    
    By the triangle inequality and the H\"older inequality, we have
    \begin{align}\label{eq:consistency__gjjung_3__ineq_main_2}
    \begin{split}
        &\|
            \gamma_t \zeta_t - \gamma^{nn}_t \zeta^{nn}_t
        \|_{l^2(\mathbb{Z}^d)}
        \\
        &\leq
        \|
            \gamma_t (\zeta_t - \zeta^{nn}_t)
        \|_{l^2(\mathbb{Z}^d)}
        +
        \|
            (\gamma_t - \gamma^{nn}_t) \zeta^{nn}_t
        \|_{l^2(\mathbb{Z}^d)}
        \\
        &\leq
        \underbrace{
            \| \gamma_t \|_{l^4(\mathbb{Z}^d)} \| \zeta_t - \zeta^{nn}_t \|_{l^4(\mathbb{Z}^d)}
        }_\text{(a)}
        +
        \underbrace{
            \| \gamma_t - \gamma^{nn}_t \|_{l^4(\mathbb{Z}^d)} \| \zeta^{nn}_t \|_{l^4(\mathbb{Z}^d)}
        }_\text{(b)}.
    \end{split}
    \end{align}
    For an upper bound of (a), since $\| \cdot \|_{l^q(\mathbb{Z}^d)} \leq \| \cdot \|_{l^p(\mathbb{Z}^d)}$ on $l^p(\mathbb{Z}^d)$ whenever $1 \leq p \leq q \leq \infty$, we have
        \begin{align*}
            \| \zeta_t - \zeta^{nn}_t \|_{l^4(\mathbb{Z}^d)}
            &\leq
            \| \zeta_t - \zeta^{nn}_t \|_{l^2(\mathbb{Z}^d)}
            \\
            &\leq
            \| (\alpha_t \hat{f} - \alpha^{nn}_t \hat{f}) * (\beta_t \hat{f} )\|_{l^2(\mathbb{Z}^d)}
            +
            \| (\alpha^{nn}_t \hat{f}) * (\beta_t \hat{f} - \beta^{nn}_t \hat{f}) \|_{l^2(\mathbb{Z}^d)}.
        \end{align*}
        Applying the convolution theorem, Parseval's identity, and the H\"older inequality, we have
        \begin{align}\label{eq:consistency__gjjung_3__ineq_main_2_part1_core}
        \begin{split}
            \| (\alpha_t \hat{f} - \alpha^{nn}_t \hat{f}) * (\beta_t \hat{f} ) \|_{l^2(\mathbb{Z}^d)}
            &=
            (2T)^{-d/2}
            \| (\alpha_t \hat{f} - \alpha^{nn}_t \hat{f})^\vee (\beta_t \hat{f} )^\vee \|_{L^2(D_T)}
            \\
            &\lesssim
            \| (\alpha_t \hat{f} - \alpha^{nn}_t \hat{f})^\vee \|_{L^4(D_T)}
            \| (\beta_t \hat{f} )^\vee \|_{L^4(D_T)},
            \\
            \| (\alpha^{nn}_t \hat{f}) * (\beta_t \hat{f} - \beta^{nn}_t \hat{f}) \|_{l^2(\mathbb{Z}^d)}
            &=
            (2T)^{-d/2}
            \| (\alpha^{nn}_t \hat{f})^\vee (\beta_t \hat{f} - \beta^{nn}_t \hat{f})^\vee \|_{L^2(D_T)}
            \\
            &\lesssim
            \| (\alpha^{nn}_t \hat{f})^\vee \|_{L^4(D_T)}
            \| (\beta_t \hat{f} - \beta^{nn}_t \hat{f})^\vee \|_{L^4(D_T)},
        \end{split}
        \end{align}
        where, for any $a \in l^1(\mathbb{Z}^d)$, $a^\vee: D_T \rightarrow \mathbb{C}$ is the function defined as $a^\vee(\boldsymbol{v}) = \sum_{k \in \mathbb{Z}^d} {a_k e^{i\frac{\pi}{T} k\cdot\boldsymbol{v}}}$ for all $\boldsymbol{v} \in D_T$.
        By Young's convolution inequality, we have
        \begin{align*}
            \| (\alpha_t \hat{f} - \alpha^{nn}_t \hat{f})^\vee \|_{L^4(D_T)}
            &=
            \| (\alpha_t - \alpha^{nn}_t)^\vee * f \|_{L^4(D_T)}
            \\
            &\leq
            \| (\alpha_t - \alpha^{nn}_t)^\vee \|_{L^2(D_T)}
            \| f \|_{L^{4/3}(D_T)}
            \\
            &\leq
            (2T)^{d/2}
            \| \alpha_t - \alpha^{nn}_t \|_{l^2(\mathbb{Z}^d)}
            \| f \|_{L^{4/3}(D_T)}
            \\
            &\lesssim
            \| \alpha_t - \alpha^{nn}_t \|_{l^2(\mathbb{Z}^d)}
            \| f \|_{L^2(D_T)}
        \end{align*}
        and similarly, we have
        $\| (\beta_t \hat{f} - \beta^{nn}_t \hat{f})^\vee \|_{L^4(D_T)}
            \lesssim
            \| \beta_t - \beta^{nn}_t \|_{l^2(\mathbb{Z}^d)}
            \| f \|_{L^2(D_T)}.$
        Furthermore,
        \begin{align*}
            \| (\beta_t \hat{f})^\vee \|_{L^4(D_T)}
            &=
            \| (\beta_t)^\vee * f \|_{L^4(D_T)}
            \leq
            \| (\beta_t)^\vee \|_{L^2(D_T)}
            \| f \|_{L^{4/3}(D_T)}
            \\
            &=
            (2T)^{-d/2}
            \| \beta_t \|_{l^2(\mathbb{Z}^d)}
            \| f \|_{L^{4/3}(D_T)}
            \lesssim
            \| \beta_t \|_{l^2(\mathbb{Z}^d)}
            \| f \|_{L^2(D_T)}
        \end{align*}
        and similarly, we have
        \begin{align*}
            \| (\alpha^{nn}_t \hat{f})^\vee \|_{L^4(D_T)}
            &\lesssim
            \| \alpha^{nn}_t \|_{l^2(\mathbb{Z}^d)}
            \| f \|_{L^2(D_T)}.
        \end{align*}
        Using these estimations, (\ref{eq:consistency__gjjung_3__ineq_main_2_part1_core}) can be estimated as
        \begin{align*}
        \begin{split}
            \| (\alpha_t \hat{f} - \alpha^{nn}_t \hat{f}) * (\beta_t \hat{f} ) \|_{l^2(\mathbb{Z}^d)}
            &\lesssim
            \| \alpha_t - \alpha^{nn}_t \|_{l^2(\mathbb{Z}^d)}
            \| \beta_t \|_{l^2(\mathbb{Z}^d)}
            \| f \|_{L^2(D_T)}^2,
            \\
            \| (\alpha^{nn}_t \hat{f}) * (\beta_t \hat{f} - \beta^{nn}_t \hat{f}) \|_{l^2(\mathbb{Z}^d)}
            &\lesssim
            \| \beta_t - \beta^{nn}_t \|_{l^2(\mathbb{Z}^d)}
            \| \alpha^{nn}_t \|_{l^2(\mathbb{Z}^d)}
            \| f \|_{L^2(D_T)}^2.
        \end{split}
        \end{align*}
        \reviewerX{
        Hence, if $0<\delta<1$ and
        \begin{align*}
            \norm{\alpha_t - \alpha^{nn}_t}_{l^2(\mathbb{Z}^d)}<\delta,
            \norm{\beta_t - \beta^{nn}_t}_{l^2(\mathbb{Z}^d)}<\delta,
            \norm{\gamma_t - \gamma^{nn}_t}_{l^2(\mathbb{Z}^d)}<\delta,
        \end{align*}
        then
        $\norm{\alpha^{nn}_t}_{l^2(\mathbb{Z}^d)} \leq \norm{\alpha_t}_{l^2(\mathbb{Z}^d)} + \norm{\alpha_t-\alpha^{nn}_t}_{l^2(\mathbb{Z}^d)} < \norm{\alpha_t}_{l^2(\mathbb{Z}^d)} + 1$, so
        \begin{align*}
            \| \gamma_t \|_{l^4(\mathbb{Z}^d)}
            \| \zeta_t - \zeta^{nn}_t \|_{l^4(\mathbb{Z}^d)}
            &\lesssim
            \| \gamma_t \|_{l^4(\mathbb{Z}^d)}
            \| \alpha_t - \alpha^{nn}_t \|_{l^2(\mathbb{Z}^d)}
            \| \beta_t \|_{l^2(\mathbb{Z}^d)}
            \| f \|_{L^2(D_T)}^2
            \\
            &\phantom{\leq}+
            \| \gamma_t \|_{l^4(\mathbb{Z}^d)}
            \| \beta_t - \beta^{nn}_t \|_{l^2(\mathbb{Z}^d)}
            \| \alpha^{nn}_t \|_{l^2(\mathbb{Z}^d)}
            \| f \|_{L^2(D_T)}^2
            \\
            &\leq
            \delta \norm{\gamma_t}_{l^4(\mathbb{Z}^d)} \left(
                \norm{\beta_t}_{l^2(\mathbb{Z}^d)}
                +
                1 + \norm{\alpha_t}_{l^2(\mathbb{Z}^d)}
            \right)
            \norm{f}_{L^2(D_T)}^2
            \\
            &=
            \delta A_t \norm{f}_{L^2(D_T)}^2.
        \end{align*}
    }
    Next, for an upper bound of (b), using Parseval's identity, H\"older inequality, Young's convolution theorem as in (a), we have
        \begin{align*}
            \| \zeta^{nn}_t \|_{l^4(\mathbb{Z}^d)}
            \leq
            \| \zeta^{nn}_t \|_{l^2(\mathbb{Z}^d)}
            &=
            (2T)^{-d/2}
            \| (\alpha^{nn}_t \hat{f})^\vee (\beta^{nn}_t \hat{f})^\vee \|_{L^2(D_T)}
            \\
            &\leq
            (2T)^{-d/2}
            \| \left(\alpha^{nn}_t\right)^\vee * f \|_{L^4(D_T)}
            \| \left(\beta^{nn}_t\right )^\vee * f \|_{L^4(D_T)}
            \\
            &\leq
            (2T)^{-d/2}
            \| \left(\alpha^{nn}_t\right)^\vee \|_{L^2(D_T)}
            \| \left(\beta^{nn}_t \right)^\vee \|_{L^2(D_T)}
            \| f \|_{L^{4/3}(D_T)}^2
            \\
            &\leq
            (2T)^{d/2}
            \| \alpha^{nn}_t \|_{l^2(\mathbb{Z}^d)}
            \| \beta^{nn}_t \|_{l^2(\mathbb{Z}^d)}
            \| f \|_{L^2(D_T)}^2,
        \end{align*}
        % so
        % \begin{align*}
        %     \| \gamma_t - \gamma^{nn}_t \|_{l^4(\mathbb{Z}^d)}
        %     \| \zeta^{nn}_t \|_{l^4(\mathbb{Z}^d)}
        %     \lesssim
        %     \| \alpha^{nn}_t \|_{l^2(\mathbb{Z}^d)}
        %     \| \beta^{nn}_t \|_{l^2(\mathbb{Z}^d)}
        %     \| \gamma_t - \gamma^{nn}_t \|_{l^4(\mathbb{Z}^d)}
        %     \| f \|_{L^2(D_T)}^2.
        % \end{align*}
    Hence, if
    $0 < \delta < 1$ and
    $\| \alpha^{nn}_t - \alpha_t \|_{l^2(\mathbb{Z}^d)} < \delta$,
    $\| \beta^{nn}_t - \beta_t \|_{l^2(\mathbb{Z}^d)}   < \delta$,
    $\| \gamma^{nn}_t - \gamma_t \|_{l^4(\mathbb{Z}^d)} < \delta$,
    then
    \begin{align}
    \label{eq:boundedness_of_neural_parameters}
    \begin{split}
        \| \beta^{nn}_t  \|_{l^2(\mathbb{Z}^d)}
        &\leq
        \| \beta^{nn}_t - \beta_t \|_{l^2(\mathbb{Z}^d)}
        +
        \| \beta_t \|_{l^2(\mathbb{Z}^d)}
        <
        1 + \| \beta_t \|_{l^2(\mathbb{Z}^d)},
        \\
        \| \gamma^{nn}_t \|_{l^4(\mathbb{Z}^d)}
        &\leq
        \| \gamma^{nn}_t - \gamma_t \|_{l^4(\mathbb{Z}^d)}
        +
        \| \gamma_t \|_{l^4(\mathbb{Z}^d)}
        <
        1 + \| \gamma_t \|_{l^4(\mathbb{Z}^d)},
    \end{split}
    \end{align}
    and it follows from (a) and (b) that
    \reviewerX{$
        \| \gamma_t \zeta_t - \gamma^{nn}_t \zeta^{nn}_t \|_{l^2(\mathbb{Z}^d)}
        \lesssim
        \delta (A_t+B_t) \| f \|_{L^2(D_T)}^2
    $}.

    By (a), (b), and the fact that $\| \cdot \|_{L^2(D_T)} \lesssim \| \cdot \|_{L^\infty(D_T)}$ on $L^\infty(D_T)$, we have
    \reviewerX{
        \begin{align*}
            &\| Q_N^{\text{fast}}(f_N,f_N) - Q^{nn}(f_N,f_N) \|_{L^2(D_T)}^2
            \\
            &\lesssim
            M
            \sum_{t=1}^{M} {
                \|
                    \gamma_t \zeta_t
                    -
                    \gamma^{nn}_t \zeta^{nn}_t
                \|_{l^2(\mathbb{Z}^d)}^2
            }
            \leq
            M^2 \delta^2 c^2 \norm{f}_{L^2(D_T)^2}^4
        \end{align*}
        with $\displaystyle{c \coloneqq \max_{1 \leq t \leq M} {(A_t + B_t)}}$.
        It follows that
        \begin{align*}
            \| Q_N^{\text{fast}}(f_N,f_N) - Q^{nn}(f_N,f_N) \|_{L^2(D_T)}
            \lesssim
            M \delta c \norm{f}_{L^2(D_T)^2}^2
            \lesssim
            M \delta c \norm{f}_{L^\infty(D_T)}^2.
        \end{align*}
        Therefore, choosing $\delta = O\left(\dfrac{\varepsilon}{1 + cM}\right)$, we have
        \begin{align*}
            \| Q_N^{\text{fast}}(f_N,f_N) - Q^{nn}(f_N,f_N) \|_{L^2(D_T)}
            \lesssim
            \varepsilon \norm{f}_{L^2(D_T)^2}^2.
        \end{align*}
    } This completes the proof.
\end{proof}
%%%%%%%%%%%%%%%%%%%%%%%%%%%%%%%%%%%%%%%%%%%%%%%%%%
We now present the proof of Proposition~\ref{prop:consistency_gjjung}, which follows by combining the results of the preceding lemmas through standard approximation and stability arguments.
%%%%%%%%%%%%%%%%%%%%%%%%%%%%%%%%%%%%%%%%%%%%%%%%%%
% Proof of the main proposition
\begin{proof}[Proof of Proposition~\ref{prop:consistency_gjjung}]
    Assume $f \in L^\infty(D_T)$. 
    By Theorem 5.1 in \cite{MR3202241},
    \begin{align*}
        \| Q(f, f) - Q_N(f_N, f_N) \|_{L^2(D_T)}
        \leq
        C \bigg(
            \| f - f_N \|_{L^2(D_T)}
            +
            \frac{\| Q(f_N, f_N) \|_{H^r(D_T)}}{N^r}
        \bigg)
    \end{align*}
    with $r \geq 0$, where $C$ is a positive real number depending on $T$ and $\| f \|_{L^2(D_T)}$. Given $\varepsilon > 0$, by Lemma \ref{lemma:consistency__gjjung_1}, there are $M \in \mathbb{N}$ and compactly supported functions $\alpha_t,\, \beta_t,\, \gamma_t: \mathbb{Z}^d \rightarrow \mathbb{C}$ for $t = 1, \cdots, M$ satisfying
    \begin{align*}
        \| Q_N(f_N,f_N) - Q_N^{\text{fast}}(f_N,f_N) \|_{L^2(D_T)}
        \leq
        \frac{\varepsilon}{2} \| f \|_{L^\infty(D_T)}^2
    \end{align*}
    for all $f \in L^\infty(D_T)$ and $N \in \mathbb{N}$.
    Also, by Lemma~\ref{lemma:consistency__gjjung_2}, there is $\delta>0$ such that
    \begin{center}
        $\| Q_N^{\text{fast}}(f_N,f_N) - Q^{nn}(f_N,f_N) \|_{L^2(D_T)} \leq \dfrac{\varepsilon}{2} \| f \|_{L^\infty(D_T)}^2$
        for all $f \in L^2(D_T)$,
    \end{center}
    whenever
    $\| \alpha^{nn}_t - \alpha_t \|_{l^2(\mathbb{Z}^d)} < \delta$,
    $\| \beta^{nn}_t - \beta_t \|_{l^2(\mathbb{Z}^d)} < \delta$, and
    $\| \gamma^{nn}_t - \gamma_t \|_{l^4(\mathbb{Z}^d)} < \delta$.
    Note that $\alpha_t,\, \beta_t,\, \gamma_t \in l^1(\mathbb{Z}^d)$ for $t = 1, \cdots, M$ since they are compactly supported.
    Hence, there are $L \in \mathbb{N}$ and $\alpha^{nn}_t,\, \beta^{nn}_t,\, \gamma^{nn}_t: \mathbb{Z}^d \rightarrow \mathbb{C}$ satisfying the following properties for $t = 1, \cdots, M$:
    \begin{enumerate}
        \item[(\romannumeral 1)]
        {
            $\alpha^{nn}_t(k) = \beta^{nn}_t(k) = \gamma^{nn}_t(k) = 0$ for all $k \in \mathbb{Z}^d$ with $\| k \|_\infty > L$.
        }
        \item[(\romannumeral 2)]
        {
            $\| \alpha^{nn}_t - \alpha_t \|_{l^2(\mathbb{Z}^d)} < \delta$,
            $\| \beta^{nn}_t - \beta_t \|_{l^2(\mathbb{Z}^d)} < \delta$,
            and
            $\| \gamma^{nn}_t - \gamma_t \|_{l^{4}(\mathbb{Z}^d)} < \delta$.
        }
    \end{enumerate}
    In particular, by the property (\romannumeral 1), we may identify $\alpha^{nn}_t$, $\beta^{nn}_t$, $\gamma^{nn}_t$ with elements in $\mathbb{C}^{N_\mathrm{trun}^d}$ with $N_\mathrm{trun} \coloneqq 2L+1$ for $t = 1, \cdots, M$.
    Finally, by the triangle inequality and Lemmas \ref{lemma:consistency__gjjung_1}-\ref{lemma:consistency__gjjung_2}, it follows that
    \begin{align*}
        \| & Q(f, f) - Q^{nn}(f_N, f_N) \|_{L^2(D_T)}
        \\ &\leq
        \| Q(f, f) - Q_N(f_N, f_N) \|_{L^2(D_T)}
        % \phantom{\leq}
        +
        \| Q_N(f_N, f_N) - Q_N^{\text{fast}}(f_N, f_N) \|_{L^2(D_T)}
        \\
        &\phantom{\leq}
        +
        \| Q_N^{\text{fast}}(f_N, f_N) - Q^{nn}(f_N, f_N) \|_{L^2(D_T)}
        \\ &\leq
        C \left(
            \| f - f_N \|_{L^2(D_T)}
            +
            \frac{\| Q(f_N, f_N) \|_{H^r(D_T)}}{N^r}
        \right)
        % &\phantom{\leq}
        +
        \frac{\varepsilon}{2} \| f \|_{L^\infty(D_T)}^2
        +
        \frac{\varepsilon}{2} \| f \|_{L^\infty(D_T)}^2
        \\
        &\leq
        C \left(
            \| f - f_N \|_{L^2(D_T)}
            +
            \frac{\| Q(f_N, f_N) \|_{H^r(D_T)}}{N^r}
        \right)
        +
        \varepsilon \| f \|_{L^\infty(D_T)}^2
    \end{align*}
    for some $M \in \mathbb{N}$, $\alpha^{nn}_t, \beta^{nn}_t, \gamma^{nn}_t \in \mathbb{C}^{N_\mathrm{trun}^d}$ and for $t = 1, \cdots, M$. Assuming further that $f \in L^\infty(D_T) \cap H^r(D_T)$, we have
    % \begin{align*}
    % \begin{split}
    %     &\| Q(f, f) - Q^{nn}(f_N, f_N) \|_{L^2(D_T)}
    %     \\
    %     &\leq
    %     \frac{C'}{N^r}\left( \| f \|_{H^r(D_T)} + \| Q(f_N, f_N) \|_{H^r(D_T)} \right)
    %     +
    %     \varepsilon \| f \|_{L^\infty(D_T)}^2
    % \end{split}
    % \end{align*}
    \begin{equation*}
        \| Q(f, f) - Q^{nn}(f_N, f_N) \|_{L^2(D_T)}\leq
        \frac{C'}{N^r}( \| f \|_{H^r(D_T)} + \| Q(f_N, f_N) \|_{H^r(D_T)} ) + \varepsilon \| f \|_{L^\infty(D_T)}^2
    \end{equation*}
    for some $C'>0$ since $\| f - f_N \|_{L^2(D_T)} \lesssim \dfrac{1}{N^r} \| f \|_{H^r(D_T)}$.
    This completes the proof.
\end{proof}
% End of the proof of the main proposition
%%%%%%%%%%%%%%%%%%%%%%%%%%%%%%%%%%%%%%%%%%%%%%%%%%

% Note that
% $
%     \| f - f_N \|_{L^2(D_T)}
%     =
%     \sum_{k \in \mathbb{Z}^d, |k|>N} |\hat{f}_k|^2
%     \rightarrow
%     0
% $
% as $N \rightarrow \infty$, since $\hat{f} \in l^2(\mathbb{Z}^d)$ as $f \in L^2(D_T)$.
% It implies that the error $ \| Q(f, f) - Q^{nn}(f_N, f_N) \|_{L^2(D_T)}$ can be made arbitrarily small.

\section{Simulation Results}\label{sec:simulation}
In this section, we evaluate the performance of the proposed FourierSpecNet framework for solving the Boltzmann equation.
The model is trained on a set of input distributions and their corresponding Boltzmann collision terms.
Once training is complete, the time evolution of the particle distribution is computed by integrating the trained neural collision operator using a third-order Runge–Kutta scheme.
We assess the effectiveness of the proposed method through a series of numerical experiments.
First, we validate the accuracy of FourierSpecNet by comparing its predictions with the analytic Bobylev–Krook–Wu (BKW)~\cite{krook1977exact} solution in the case of Maxwellian molecules.
Next, we evaluate the approximation quality of the learned collision operator by comparing it with classical spectral solvers in two-dimensional settings, including hard-sphere and inelastic collisions.
Finally, we investigate the performance of the method in a high-dimensional scenario by considering the three-dimensional case.
Unless otherwise specified, all experiments are conducted using a fixed model configuration with $N_{\text{trun}} = 8$ and $M = 2$.
The network is trained using the Adam optimizer~\cite{kingma2014adam} with a learning rate of $10^{-2}$, and training is terminated once the relative $L^2$ training error falls below $10^{-2}$. \reviewerY{Here, the training loss denotes the relative $L^2$ error of the collision operator $Q(f,f)$, whereas the testing errors in the following figures correspond to the relative $L^2$ error of the distribution function $f$ obtained after time integration. Thus, an operator-level loss of order $10^{-2}$ typically results in a solution-level error of $10^{-2}$–$10^{-3}$ owing to the stability of the time integration.}
\reviewerX{Moreover, setting a stricter tolerance (e.g., \(10^{-3}\)) would reduce the propagated error but lead to increase in training time. Therefore, we chose \(10^{-2}\) as the optimal early stopping criterion, balancing both stability of the simulation and computational efficiency of the proposed method.}
\reviewerZ{Throughout all experiments, we provide tables which compare the runtime of the fast spectral method and FourierSpecNet to find the solution of the Boltzmann equation for various grid resolutions ($N = 2^k$ with $k \in \{4, 5, 6, 7, 8, 9\}$). For a fair comparision, both methods were implemented and executed with an A100 GPU on PyTorch 2.5.1. The epochs and training times for the experiments in Sections 4.1-4.3 are summarized in Table~\ref{tab:models_training_time}.}

\begin{table}[h]
\reviewerZ{
    \centering
    \resizebox{\linewidth}{!}{
        \begin{tabular}{c|ccc}
\hline
Experiment      &   Maxwellian (Sec. 4.1)  &   Hard sphere (Sec. 4.2)     &   Inelastic (Sec. 4.3)\\
\hline
Epochs          &   $2 \times 10^5$     &   $2 \times 10^5$   &    $5 \times 10^5$\\
Training time   &   01h 39m     &   01h 42m   &    04h 19m\\
\hline
\end{tabular}
    }
    \caption{Summary of the epochs and training time for the experiments in Sections 4.1-4.3.}
    \label{tab:models_training_time}
}
\end{table}

\subsection{Maxwellian molecules case}
%%%%% Time-evolution
\begin{figure}[t]
\begin{center}
\includegraphics[width=\textwidth]{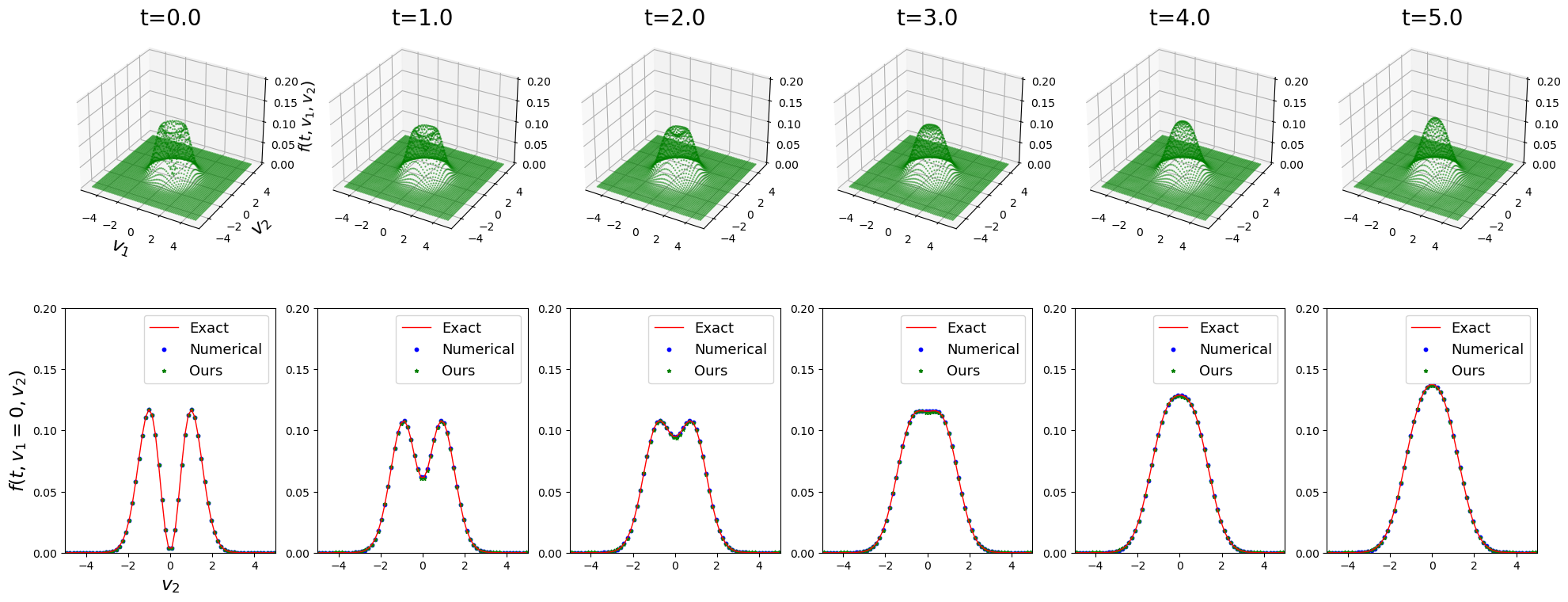}
\end{center}
\caption{Time evolution of the BKW solution for Maxwellian molecules. The top row shows the pointwise predicted values of $f^{nn}(t, \boldsymbol{v})$ (FourierSpecNet) over time ($t = 0$ to $t = 5$). The bottom row illustrates detailed 2D slices of the velocity space at specific time instances, comparing FourierSpecNet predictions with the exact analytical solution (dotted lines) and numerical results (solid lines).}
\label{fig:max_trajectory}
\end{figure}
%%%%% Conservation laws
%%%%% Zero-shot super-resolution and error curves
\begin{figure}[t]
\begin{center}
\includegraphics[width=\textwidth]{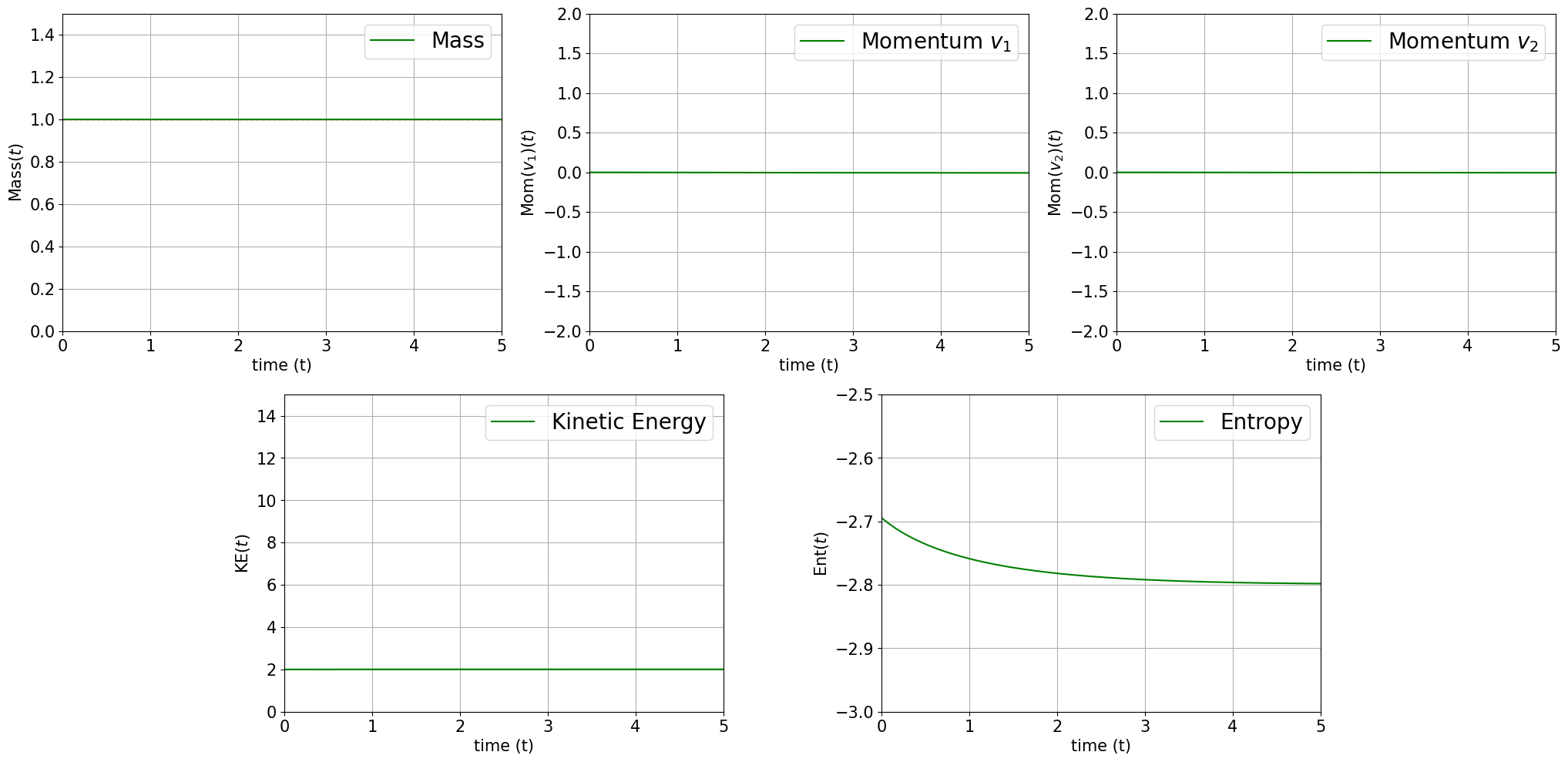}
\end{center}
\caption{Physical quantities of the BKW solution ($f^{nn}$) as a function of time ($t = 0$ to $t = 5$). The predicted values (green lines) for mass, momentum ($v_1$ and $v_2$), kinetic energy, and entropy are plotted.}
\label{fig:max_physical}
\end{figure}
%%%%% Zero-shot super-resolution and error curves
\begin{figure}[t]
\begin{center}
\includegraphics[width=\textwidth]{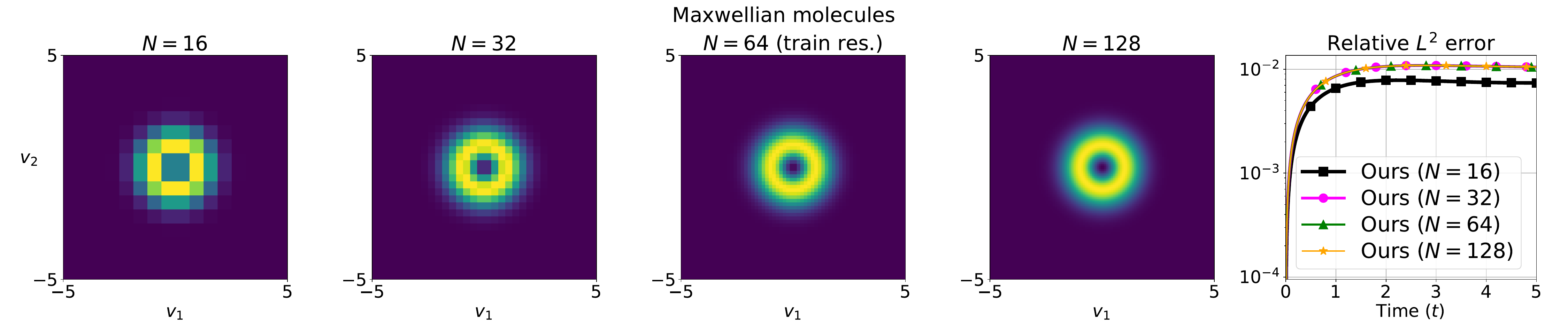}   % In the revised manuscript
\end{center}
\caption{Evaluation of the resolution-invariance and super-resolution capability of FourierSpecNet. The left panel shows the predicted collision operator $Q(f, f)$ for various grid resolutions ($N = 16, 32, 64, 128$), demonstrating the model's ability to generalize across different resolutions. The right panel presents the relative $L^2$-error of FourierSpecNet (denoted as 
`Ours') and the spectral method over time ($t = 0$ to $t = 5$), highlighting FourierSpecNet's consistent accuracy across all resolutions.}
\label{fig:max_super}
\end{figure}

The BKW solution, a well-known analytic solution for Maxwellian molecules, serves as a benchmark to evaluate the accuracy of numerical methods applied to the Boltzmann equation. The solution is given as
\begin{equation}
  f(t,v)=\frac{1}{2\pi S^2}e^{-\frac{|v|^2}{2S\sigma^2}}\left(2S-1+\frac{1-S}{2S}\frac{|v|^2}{\sigma^2}\right),
\end{equation}
where $S=1-\frac{1}{2}e^{-\frac{t}{8}}$. We assessed the accuracy and convergence of FourierSpecNet by comparing between the predicted and exact solutions. For time evolution, a third-order Runge-Kutta method was employed with a final time $t_\text{final}=5$ and a time step $dt = 0.01$. Figure \ref{fig:max_trajectory} shows the predicted and exact solutions over time, illustrating FourierSpecNet’s ability to accurately approximate the BKW solution dynamics. The model consistently achieves high accuracy throughout the simulation.

To further validate FourierSpecNet, we examined its preservation of physical moments, such as density, momentum, and energy defined in \eqref{eq:physics_quantity}, which are critical in kinetic theory. These quantities were computed at each time step from  $t = 0$  to  $t = 5$. As shown in Figure \ref{fig:max_physical}, the predicted moments closely match the exact values, demonstrating that FourierSpecNet preserves these physical properties, a key requirement for kinetic simulations.

We also assessed the resolution-invariance of FourierSpecNet by evaluating its super-resolution capability across different grid resolutions ($N = 16, 32, 64, 128$). The model is trained using data generated at a fixed resolution of $N = 64$ and tested on inputs at varying resolutions without retraining. As shown in Figure~\ref{fig:max_super}, the relative $L^2$-error remains consistently low across all resolutions, demonstrating that FourierSpecNet generalizes well to unseen resolutions. These results underscore the model’s efficiency and its applicability to large-scale, high-dimensional kinetic simulations.

\reviewerZ{
    Figure~\ref{fig:max_super} shows that the model trained at $N=64$ exhibits a slightly lower error at $N=16$ while exhibiting similar level of error at $N \geq 32$.
    This discrepancy can be explained by the spectral characteristics of the input representation rather than an architectural limitation.
    Specifically, the data at $N=16$ are heavily downsampled from the original distribution, leading to the loss of high-frequency components and aliasing in the spectral domain.
    Consequently, the predicted solutions at $N=16$ appear smoother, and the relative $L^2$ error becomes smaller because fine-scale discrepancies are averaged out rather than faithfully resolved.
    This effect is thus a numerical artifact of spectral truncation rather than a true improvement in physical accuracy.
    In contrast, when the resolution is equal to or higher than the training resolution ($N \geq 64$), the model operates within the same spectral band as during training, leading to consistent performance across resolutions.
    This demonstrates that FourierSpecNet achieves resolution invariance within the trained frequency range, validating its zero-shot generalization capability in the spectral domain.
}

\reviewerZ{
    Finally, we also assessed the efficiency of FourierSpecNet by measuring the time to compute the solution of the elastic Boltzmann equation across different grid resolutions.
    As shown in Table~\ref{tab:inference_time__maxwellian}, FourierSpecNet computes faster than the fast spectral method with increasing ratio of the computation time in the resolution.
    This result shows that FourierSpecNet is computationally more efficient than the fast spectral method to compute the solution of the Boltzmann equation.
}

\begin{table}[h] % Tabel begins
\reviewerZ{
    \caption{
        Comparison of inference times for the fast spectral method and FourierSpecNet at $t = 5.0$ with $N = 2^k\, (k \in \{4, 5, 6, 7, 8, 9\})$ for the Maxwellian molecules. 
        The classical RK3 method with $\Delta t = 0.01$ was used for time discretization.
        % After training, FourierSpecNet demonstrates consistent efficiency across resolutions, significantly reducing inference time compared to the fast spectral method.
    }
}
\begin{center}
\reviewerZ{\begin{tabular}{c|cccccc}
\hline
$N$ & 16 & 32 & 64 & 128 & 256 & 512\\
\hline
Fast spectral method (\(t_1\)) & 0.40s & 0.52s & 1.57s & 6.02s & 22.52s & 90.68s\\
(Ours) FourierSpecNet (\(t_2\)) & 0.51s & 0.52s & 0.59s & 0.57s & 0.72s & 1.31s\\
\hline
Speed-up (\(t_1/t_2\)) & $\times$0.78 & $\times$1.00 & $\times$2.65 & $\times$10.59 & $\times$31.20 & $\times$69.02\\
\hline
\end{tabular}}
\end{center}

\label{tab:inference_time__maxwellian}
\end{table} % Tabel ends

\subsection{Hard sphere molecules case}
%%%%% Time-evolution
\begin{figure}[t]
\begin{center}
\includegraphics[width=\textwidth]{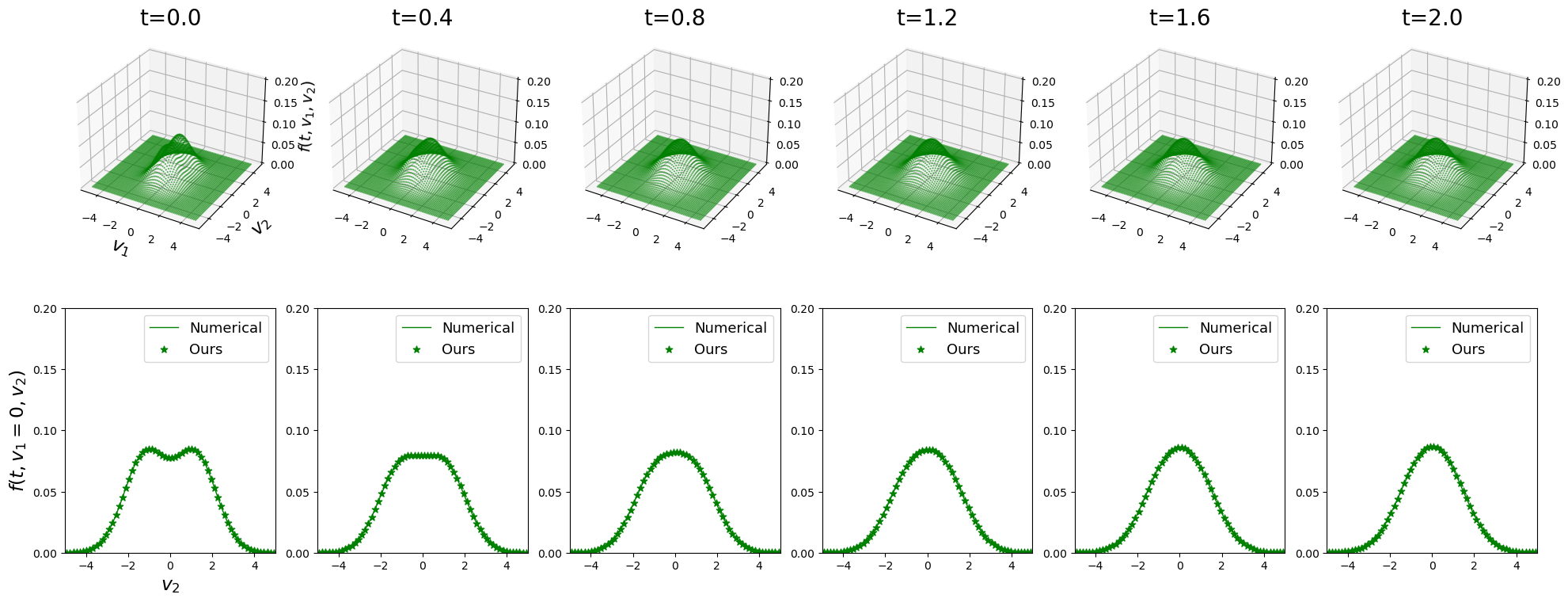}
\end{center}
\caption{Time evolution of the solution for hard-sphere molecules. The top row shows 2D slices of the velocity space predicted by FourierSpecNet at various time steps ($t = 0, 0.4, 0.8, 1.2, 1.6, 2.0$), and the bottom row provides pointwise comparisons of $f^{nn}(t, v)$ with the numerical results obtained using the spectral method.}
\label{fig:hard_trajectory}
\end{figure}
%%%%% Conservation laws
\begin{figure}[t]
\begin{center}
\includegraphics[width=\textwidth]{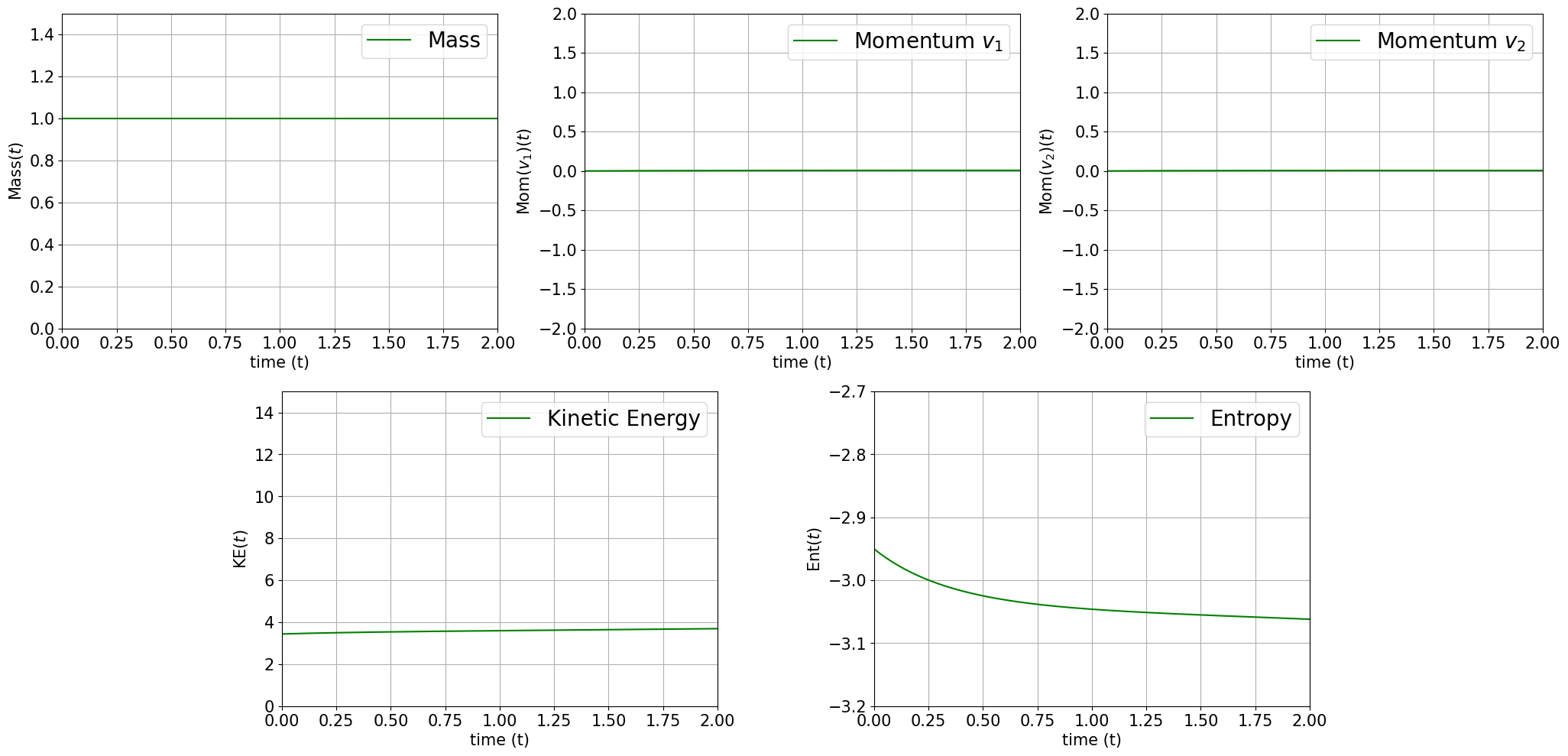}
\end{center}
\caption{Physical quantities of the solution for hard sphere molecules ($f^{nn}$) over time ($t = 0$ to $t = 2$). The predicted values (green lines) for mass, momentum ($v_1$ and $v_2$), kinetic energy, and entropy are shown.}
\label{fig:hard_physical}
\end{figure}
%%%%% Zero-shot super-resolution and error curves
\begin{figure}[t]
\begin{center}
\includegraphics[width=\textwidth]{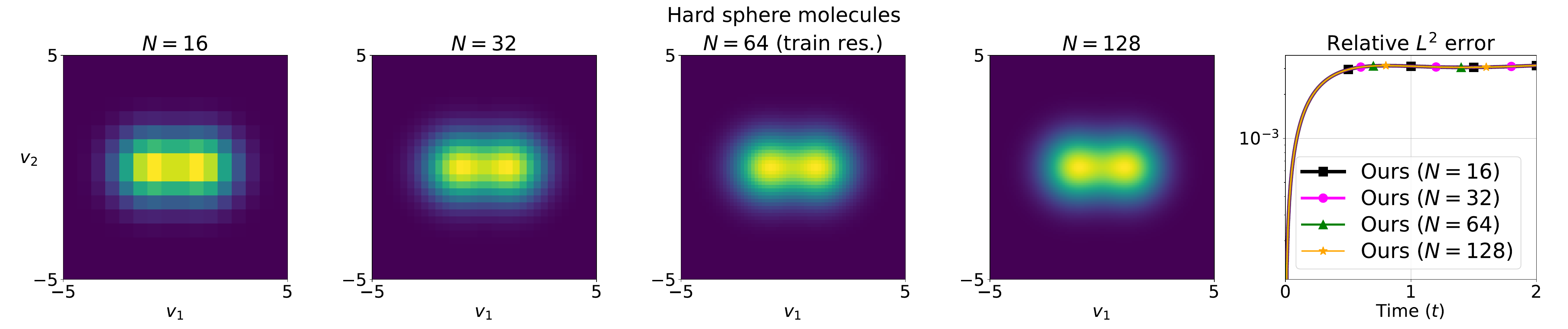}
\end{center}
\caption{Evaluation of the resolution-invariance and super-resolution capability of FourierSpecNet for hard-sphere molecules. The left panel shows the predicted collision operator $Q(f, f)$ for various grid resolutions ($N = 16, 32, 64, 128$), demonstrating the model's ability to generalize across resolutions. The right panel presents the relative $L^2$-error between the FourierSpecNet predictions (denoted as `Ours') and the fast spectral method over time ($t = 0$ to $t = 2$), highlighting consistent accuracy across all resolutions. These results confirm the resolution-invariance property of FourierSpecNet.}
\label{fig:hard_super}
\end{figure}

In the case of hard sphere molecules, there is no well-known exact solution to the Boltzmann equation. Therefore, we evaluate the performance of FourierSpecNet by comparing the Boltzmann collision term obtained using the fast spectral method proposed in~\cite{MR3687853} and our proposed approach. The initial condition for this test case is given by:
\begin{equation}
    f_{\text{init}}(\boldsymbol{v}) =C\bigg( \frac{1}{2\pi\sigma_1^2}e^{-\frac{|\boldsymbol{v}-\mu_1|^2}{2\sigma_1^2}} + \frac{1}{2\pi\sigma_2^2}e^{-\frac{|\boldsymbol{v}-\mu_2|^2}{2\sigma_2^2}}\bigg),
\end{equation}
where $\mu_1=(0,-1.2),\mu_2=(0,1.2),\sigma_1=\sigma_2=1$ and $C$ is the normalizing constant to make the volume of the distribution 1.

After training, we compared the trajectory of the solution predicted by FourierSpecNet with the results obtained using the fast spectral method. Figure \ref{fig:hard_trajectory} shows the time evolution of the solutions over $t_\text{final}=2$. The FourierSpecNet predictions closely follow the trajectory of the fast spectral method results, demonstrating the model’s ability to approximate the collision operator with high accuracy. This highlights the robustness of FourierSpecNet even in the absence of a known exact solution.

To further validate the model, we examined the physical moments of the predicted solution, including density, momentum, and energy. These quantities were computed at every time step from $t=0$ to $t=2$. As shown in Figure \ref{fig:hard_physical}, FourierSpecNet successfully preserves these key physical properties, which are crucial for ensuring the reliability of kinetic simulations.

In addition to accuracy, we evaluated the super-resolution capability of FourierSpecNet. Figure \ref{fig:hard_super} illustrates the relative $L^2$-error between the predicted and reference solutions across different grid resolutions $(N=16, 32, 64, 128)$. The model is also trained using data generated at a fixed resolution of $N = 64$ and tested on inputs at varying resolutions without retraining. The results demonstrate that the $L^2$-error remains consistent regardless of the grid resolution, confirming the resolution-invariance property of FourierSpecNet. This capability allows the model to perform accurate high-resolution inference even when trained at lower resolutions, significantly reducing computational costs.

Finally, we compared the inference time of FourierSpecNet with that of the fast spectral method.
As summarized in Table~\ref{tab:inference_time__hard_sphere}, FourierSpecNet achieves significantly faster inference times while maintaining competitive accuracy.
\reviewerZ{Although FourierSpecNet requires a pre-training phase, this training can be efficiently performed on modern GPUs, whose computational power significantly reduces the learning time. Once trained, the model encodes the collision operator into a set of learnable spectral coefficients supported on a fixed truncated frequency band, whose parameter count does not scale with the grid resolution $N$. Consequently, its inference cost remains essentially constant even as $N$ increases, whereas the classical fast spectral method must repeatedly perform quadrature-based convolution operations at each resolution level.} This improvement thus primarily stems from the pretrained operator representation and the resolution-invariant structure of FourierSpecNet, \reviewerZ{rather than from GPU acceleration alone. It highlights that the FourierSpecNet has comparable or faster inference than the fast spectral method across resolutions, while maintaining similar accuracy.}
These results highlight the computational efficiency of FourierSpecNet, making it a viable alternative to the spectral methods, particularly for large-scale simulations.

\begin{table}[t] % Tabel begins
\caption{
    Comparison of inference times for the fast spectral method and FourierSpecNet at \reviewerTWO{$t = 5.0$} with $N = 2^k\, (k \in \{4, 5, 6, 7, 8, 9\})$ \reviewerTWO{for the elastic hard sphere molecules}.
    The classical RK3 method with $\Delta t = 0.01$ was used for time discretization. 
    % \reviewerZ{Both methods are executed on the same GPU hardware for a fair comparison. 
    % % After training, FourierSpecNet demonstrates consistent efficiency across resolutions, significantly reducing inference time compared to the fast spectral method.
    % }
}
\begin{center}
\begin{tabular}{c|cccccc}
\hline
$N$ & 16 & 32 & 64 & 128 & 256 & 512\\ \hline
Fast spectral method (\(t_1\)) & 0.39s & 0.51s & 1.57s & 5.16s & 22.61s & 90.81s\\
(Ours) FourierSpecNet (\(t_2\)) & 0.51s & 0.53s & 0.53s & 0.66s & 0.74s & 1.26s\\
\hline
Speed-up (\(t_1/t_2\))& $\times$0.77 & $\times$0.98 & $\times$2.97 & $\times$7.88 & $\times$30.71 & $\times$71.99\\
\hline
\end{tabular}
\end{center}
\label{tab:inference_time__hard_sphere}
\end{table} % Tabel ends
%%%%%%%%%%%%%%%%%%%%%%%%%%%%%%%%%%%%%%%%%%%%%%%%%%
%%%%%%%%%%%%%%%%%%%%%%%%%%%%%%%%%%%%%%%%%%%%%%%%%%
% % The first table is on the arXiv version.
% \begin{center}
% \begin{tabular}{ccccc}
% \hline
% $N$ & 16 & 32 & 64 & 128 \\ \hline
% Fast spectral method & 5.74 (s) & 22.71 (s) & 86.87(s) & 251.50(s) \\
% FourierSpecNet (Ours) & 0.62 (s) & 0.77(s) & 0.63(s) & 0.81(s) \\ \hline
% \end{tabular}
% \end{center}
% %%%%%%%%%%%%%%%%%%%%

\subsection{Inelastic Boltzmann case}
%%%%% Time-evolution
\begin{figure}[t]
\begin{center}
\includegraphics[width=\textwidth]{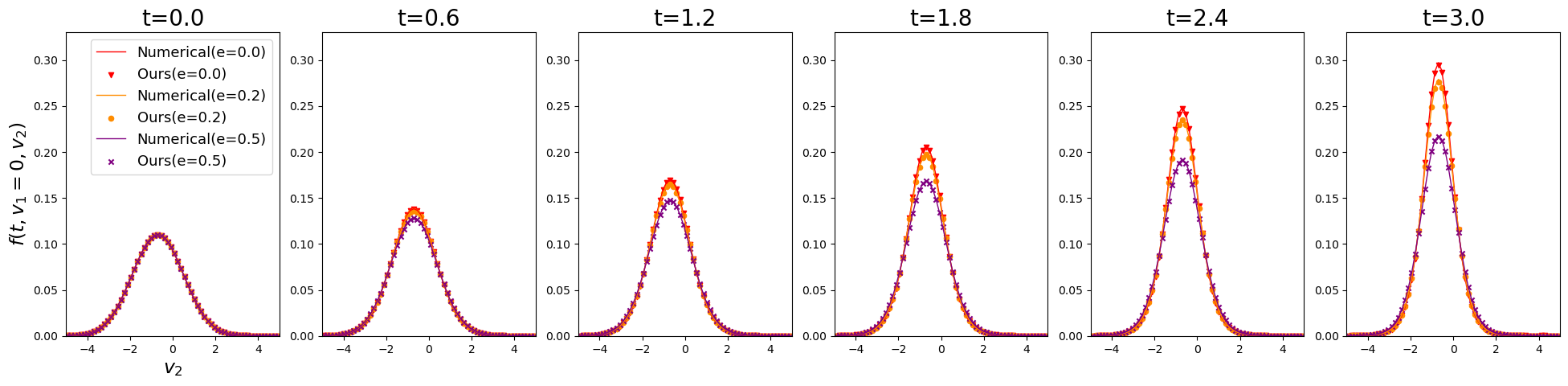}
\includegraphics[width=\textwidth]
{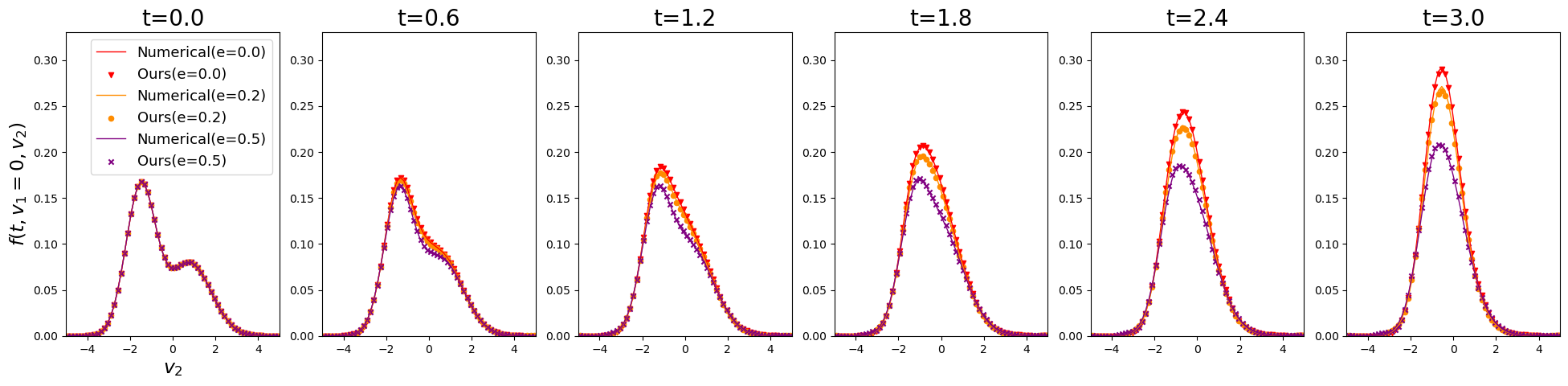}
\end{center}
\caption{Time evolution of the inelastic Boltzmann equation solution $f^{nn}(t, v)$ for different elasticity parameters ($e = 0.0, 0.2, 0.5$) under two distinct initial conditions (Upper: \eqref{eq:init_inela_1}, Lower: \eqref{eq:init_inela_2}). The predicted results from FourierSpecNet are compared with numerical solutions obtained using the spectral method at various time steps ($t = 0, 0.6, 1.2, 1.8, 2.4, 3.0$). The results highlight FourierSpecNet’s ability to accurately approximate the collision operator across different elasticity scales.}
\label{fig:inelastic_trajectory}
\end{figure}
%%%%% Conservation laws
% None
%%%%% Zero-shot super-resolution and error curves
\begin{figure}[t]
\begin{center}
\includegraphics[width=\textwidth]{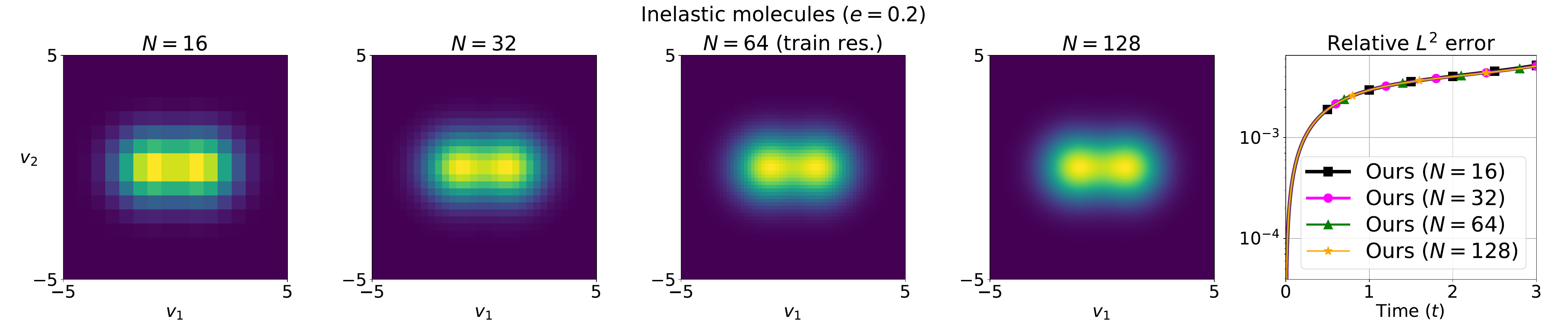}
\end{center}
\caption{Evaluation of the relative $L^2$-error for FourierSpecNet and the fast spectral method across different grid resolutions ($N = 16, 32, 64, 128$) for the inelastic Boltzmann equation with initial condition \eqref{eq:init_inela_2}. The left panel illustrates the predicted collision operator $Q(f, f)$ for each resolution, while the right panel shows the relative $L^2$-error over time ($t = 0$ to $t = 3$). The results demonstrate FourierSpecNet’s ability to achieve high accuracy across various resolutions, even when trained at lower resolutions.}
\label{fig:inelastic_super}
\end{figure}

To further evaluate the versatility of FourierSpecNet, we applied it to the inelastic Boltzmann equation, which introduces additional challenges due to the energy dissipation during collisions. In this case, we compare between the predictions of FourierSpecNet and the numerical results obtained using the fast spectral method proposed in~\cite{MR3687853}. The following two initial conditions are considered:
\begin{equation}\label{eq:init_inela_1}
f^{(1)}_{\text{init}}(\boldsymbol{v}) = C\frac{1}{2\pi\sigma^2}e^{-\frac{|\boldsymbol{v}-\mu|^2}{2\sigma^2}},
\end{equation}
where $\mu=(0.0,-0.7)$ and $\sigma=1.2$, and
\begin{equation}\label{eq:init_inela_2}
f^{(2)}_{\text{init}}(\boldsymbol{v}) = C \bigg( \frac{1}{2\pi\sigma_1^2}e^{-\frac{|\boldsymbol{v}-\mu_1|^2}{2\sigma_1^2}} + \frac{1}{2\pi\sigma_2^2}e^{-\frac{|\boldsymbol{v}-\mu_2|^2}{2\sigma_2^2}} \bigg),
\end{equation}
where $\mu_1 = (0.0, -1.5)$, $\mu_2 = (0.0, 0.8)$, $\sigma_1 = 0.7$, and $\sigma_2 = 1$. In both cases, the constant $C$ denotes the normalization factor that ensures the total integral of the distribution equals 1. Figure \ref{fig:inelastic_trajectory} illustrate the evolution of the numerical solution and FourierSpecNet predictions over time ($t = 0$ to $t = 3$) for each two different initial conditions. The results demonstrate that FourierSpecNet maintains a consistently low error throughout the simulation. This indicates that the proposed method effectively approximates the collision operator even in the presence of energy dissipation, a key characteristic of inelastic collisions. FourierSpecNet's ability to handle this additional complexity highlights its adaptability to diverse kinetic problems.

% \begin{figure}
%     \centering
%     \subfigure{\includegraphics[width=0.4\textwidth]{images/rel_L2_error_elastic0.0.png}}

%     \caption{Time evolution of $ \| f_{num} - f_{pred} \|_{L^2(\mathbb{R}^2)} $ for $N=32,64,128$. $f_{num}$ is a solution obtained by spectral method in (bib) and $f_{pred}$ is by our approach. Classical RK3 with $ \Delta t = 0.01 $ for time discretization.}
% \end{figure}

In addition, the resolution-invariance of FourierSpecNet was tested by examining its super-resolution capability. Figure \ref{fig:inelastic_super} shows the $L^2$-error across various grid resolutions ($N = 16, 32, 64, 128$). The results reveal that the $L^2$-error remains stable regardless of the grid size, confirming that FourierSpecNet can generalize across resolutions without retraining. This property is particularly advantageous for computationally expensive problems, as it enables high-resolution inference while significantly reducing the training and computation costs. These results underscore the robustness and flexibility of FourierSpecNet in approximating the inelastic Boltzmann equation.

%%%%%%%%%%%%%%%%%%%%%%%%%%%%%%%%%%%%%%%%%%%%%%%%%%
%%%%%%%%%%%%%%%%%%%%%%%%%%%%%%%%%%%%%%%%%%%%%%%%%%
% Table of the elapsed time for the Maxwellian molecules
\begin{table}[] % Tabel begins

\begin{center}
\reviewerZ{
    \caption{
    Comparison of inference times for the fast spectral method and FourierSpecNet at $t = 5.0$ with $N = 16, 32, 64, 128, 256$ for the inelastic Maxwellian molecules of the elasticity $e=0.5$.
    }
    \begin{tabular}{c|cccccc}
\hline
$N$ & 16 & 32 & 64 & 128 & 256 & 512\\
\hline
Fast spectral method (\(t_1\)) & 0.42s & 0.51s & 1.66s & 5.16s & 22.61s & 90.79s\\
(Ours) FourierSpecNet (\(t_2\)) & 0.55s & 0.56s & 0.58s & 0.61s & 0.78s & 1.26s\\
\hline
Speed-up (\(t_1/t_2\)) & $\times$0.76 & $\times$0.91 & $\times$2.88 & $\times$8.40 & $\times$29.08 & $\times$71.95\\
\hline
\end{tabular}
}
\end{center}

\label{tab:inference_time__inelastic}
\end{table} % Tabel ends
%%%%%%%%%%%%%%%%%%%%%%%%%%%%%%%%%%%%%%%%%%%%%%%%%%
%%%%%%%%%%%%%%%%%%%%%%%%%%%%%%%%%%%%%%%%%%%%%%%%%%

\subsection{Three dimensional in velocity case}
\begin{figure}[t]
\begin{center}
\includegraphics[width=0.7\textwidth]{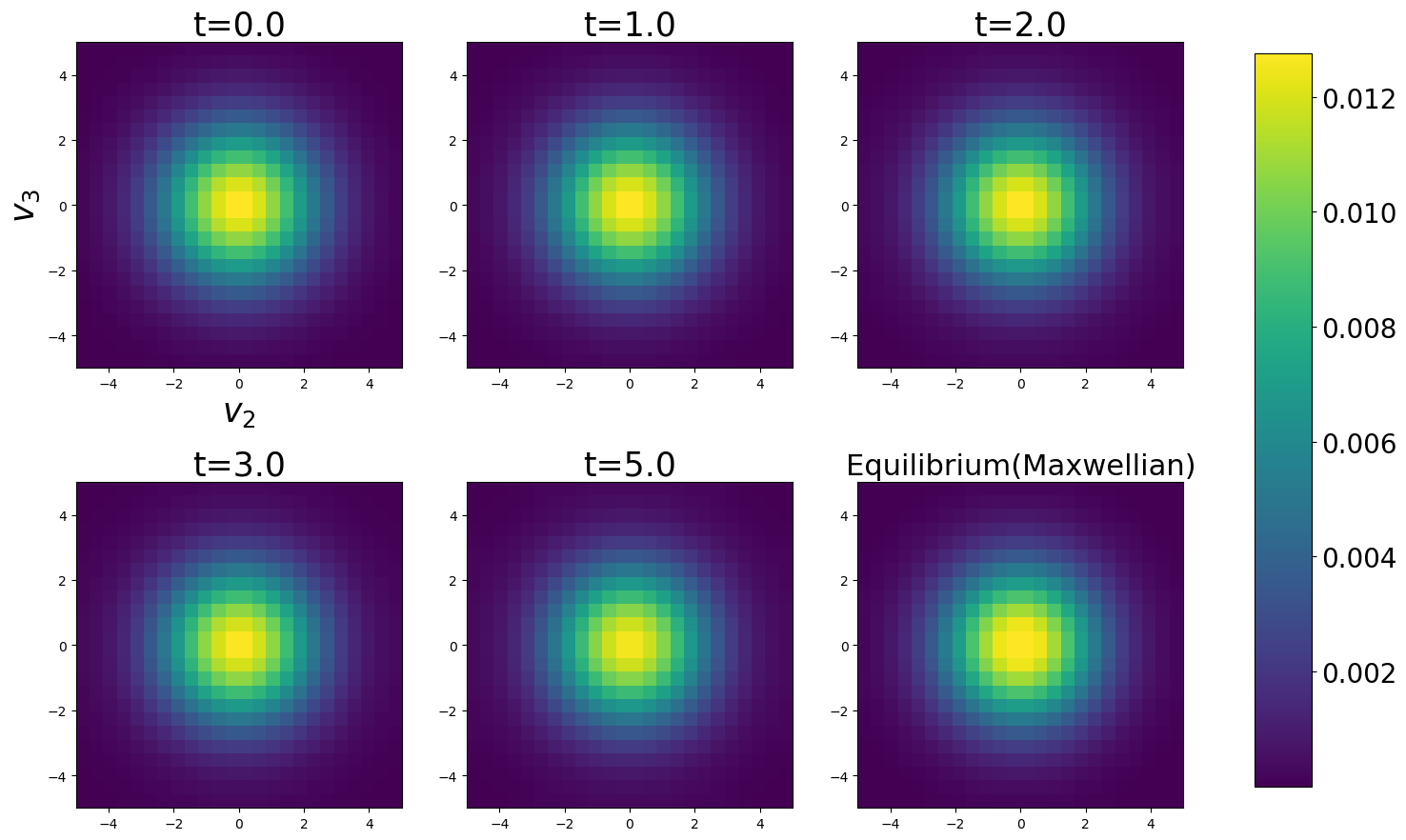}
\end{center}
\caption{Evolution of the Maxwellian initial distribution in a three-dimensional velocity space ($v_1$, $v_2$, $v_3$) predicted by FourierSpecNet at different time steps ($t = 0, 1.0, 2.0, 3.0, 5.0$) and at equilibrium (Maxwellian). The color scale represents the relative $L^2$-error between FourierSpecNet predictions and the fast spectral method.}
\label{fig:3d}
\end{figure}
To further evaluate the performance and scalability of FourierSpecNet, we extend our experiments to a three-dimensional velocity space. This case represents a significant computational challenge due to the increased dimensionality of the Boltzmann collision operator. The fast spectral methods face scalability issues in such settings, as the computational cost grows exponentially with the number of discretization points in each dimension.

In this experiment, we use a Maxwellian initial distribution as the baseline condition and compute the collision operator over time using FourierSpecNet. The model is trained on lower-resolution data but is tasked with performing high-resolution inference, highlighting its zero-shot super-resolution capability. The accuracy of the collision operator is assessed by comparing FourierSpecNet’s predictions with the results obtained from the spectral method. Figure \ref{fig:3d} illustrates the relative  $L^2-$error of FourierSpecNet compared to the fast spectral method across different grid resolutions. The results demonstrate that FourierSpecNet maintains high accuracy and stability across a wide range of resolutions, even as the dimensionality of the velocity space increases. This consistency in performance showcases the robustness of FourierSpecNet and its ability to generalize effectively in higher-dimensional settings. Moreover, the computational efficiency of the proposed method remains significantly higher than traditional approaches, as FourierSpecNet leverages GPU acceleration for rapid batch inference. These results emphasize the scalability and adaptability of FourierSpecNet for solving the Boltzmann equation in complex, high-dimensional scenarios. FourierSpecNet demonstrates its potential as a versatile tool for advancing the numerical simulation of kinetic equations.

\section{Conclusion}\label{sec:conclusion}
In this work, we introduced FourierSpecNet, a novel deep learning based framework for approximating solutions to the Boltzmann equation. By incorporating deep learning to optimize the convolution process in the fast spectral method, FourierSpecNet achieves significant improvements in computational efficiency while maintaining the accuracy and stability of traditional numerical methods. Through extensive experiments, we demonstrated that FourierSpecNet effectively captures the dynamics of the Boltzmann equation and preserves key physical moments such as density, momentum, and energy. FourierSpecNet was validated across diverse scenarios, including Maxwellian molecules, hard sphere collisions, and inelastic cases. The model consistently achieved low relative errors and demonstrated its resolution-invariance, enabling accurate high-resolution predictions from low-resolution training data. FourierSpecNet also admits a theoretical consistency guarantee \reviewerX{as in Proposition \ref{prop:consistency_gjjung}}, which establishes that the approximation error of the trained collision operator can be bounded by the spectral truncation error and vanishes as the resolution increases. These results highlight the versatility of FourierSpecNet and its potential to address a broad range of kinetic problems.

Future research will focus on extending FourierSpecNet to more complex settings, such as Boltzmann equations with boundary conditions. Traditional numerical methods often struggle with boundary interactions due to the intricate coupling between boundary effects and interior dynamics. By integrating FourierSpecNet with advanced architectures like graph neural networks (GNNs), we aim to develop a hybrid framework that efficiently models Boltzmann equations in a bounded domain while maintaining the scalability and adaptability demonstrated in this work. This extension would pave the way for solving real-world kinetic problems with greater complexity.

\appendix
\section{Decay of the kernel modes for the VHS model}
\label{sec:decay_of_kernel_modes}

This section aims to investigate the decay of the kernel modes $\hat{G}(l, m)$ as $\| l \|_\infty + \| m \|_\infty \rightarrow \infty$ for the variable hard sphere (VHS) model. More specifically, this section provides a proof of the following lemma.
\begin{lemma}[Decay of the kernel modes]
    \label{lemma:decay of the kernel modes}
    Let $B$ be the collision kernel of the variable hard sphere (VHS) model with $\alpha>-1$, and assume $e \in [0, 1] \cap \mathbb{Q}$.
    For each $\epsilon > 0$, there is $K>0$ for which $\left| \hat{G}(l, m) \right| < \epsilon$ whenever $l, m \in \mathbb{Z}^d$ and $\| l \|_\infty > K$ or $\| m \|_\infty > K$.
\end{lemma}
For the VHS model, we have
\begin{align*}
    &\hat{G}(l, m)
    =
    C(2\pi)^2(2\lambda T)^{2+\alpha}
    \int_0^1 {x^{1+\alpha} J_0(\eta x) J_0(\xi x)}\, dx
    &(d=2),
    \\
    &\hat{G}(l, m)
    =
    C(4\pi)^2(2\lambda T)^{3+\alpha}
    \int_0^1 {x^{1+\alpha} \mathrm{sinc}(\eta x) \mathrm{sinc}(\xi x)}\, dx
    &(d=3),
\end{align*}
where $\lambda \coloneqq \dfrac{2}{3 + \sqrt{2}}$, $\eta \coloneqq \left\| \dfrac{1+e}{2}l - \dfrac{3-e}{2}m \right\|_2 \lambda \pi$ and $\xi \coloneqq \dfrac{1+e}{2} \left\| l+m \right\|_2 \lambda\pi$, $J_0$ is the Bessel function of order 0, and $\mathrm{sinc}(x) = \sin(x) / x$ for all $x \in \mathbb{R} \setminus \{0\}$ and $\mathrm{sinc}(0) = 1$. In the proof of Lemma \ref{lemma:decay of the kernel modes}, we utilize the following lemmas. The first lemma is a direct corollary of Equation (9.2.1) in \cite{math_functions}.

\begin{lemma}
\label{lemma:appendix_estimation_bessel_0}
    Let $J_0: \mathbb{R} \rightarrow \mathbb{R}$ be the Bessel function of order 0.
    Then
    \begin{align*}
        \lim_{x \rightarrow \infty}{
            \frac{
                J_0(x)
            }{
                \sqrt{\frac{2}{\pi x}} \cos\left( x - \frac{\pi}{4} \right)
            }
        }
        =
        1.
    \end{align*}
    In particular, $|J_0(x)| \lesssim x^{-1/2}$ for $x \in (0, \infty)$.
\end{lemma}
% \begin{proof}
%     Assume $z \in \mathbb{C}$
%     From equation (9.2.1) in \cite{math_functions}, as $|z| \rightarrow \infty$, we have
%     \begin{align*}
%         J_\nu(z) = \sqrt{\frac{2}{\pi z}} \left\{
%             \cos\left( z - \frac{\nu\pi}{2} - \frac{\pi}{4} \right)
%             +
%             \exp\left( \left| \mathrm{Im}(z) \right| \right) O(|z|^{-1})
%         \right\}
%         \quad
%         (| \arg(z) | < \pi),
%     \end{align*}
%     where $J_\nu$ is the Bessel function of order $\nu$ and $\mathrm{Im}(z)$ denotes the imaginary part of a complex number $z$.
%     With $\nu = 0$ and $x \in (0, \infty)$, it follows that $J_0(x) = \sqrt{\dfrac{2}{\pi x}} \cos\left( x - \dfrac{\pi}{4} \right) + O(x^{-3/2})$ as $x \rightarrow \infty$.
% \end{proof}

\begin{lemma}
\label{lemma:appendix_bounded_subset}
    For each $a > 0$, define
    \begin{align*}
        S_1(a)
        &\coloneqq
        \left\{
            (l, m) \in \mathbb{Z}^d \times \mathbb{Z}^d
            :
            \eta \leq a,\, \xi \leq a
        \right\},\\
        S_2(a)
        &\coloneqq
        \left\{
            (l, m) \in \mathbb{Z}^d \times \mathbb{Z}^d
            :
            \eta \leq a,\, \xi = 0
        \right\},\\
        S_3(a)
        &\coloneqq
        \left\{
            (l, m) \in \mathbb{Z}^d \times \mathbb{Z}^d
            :
            \eta = 0,\, \xi \leq a
        \right\}.
    \end{align*}
    Then, for any $a>0$, there is $R>0$ such that $\| l \|_\infty < R$ and $\| m \|_\infty < R$ whenever $(l, m) \in S_i(a)$ for $i=1, 2, 3$.
\end{lemma}
\begin{proof}
    Assume $(l, m) \in S_1(a)$.
    Then $\| u \|_2 \leq c_1$ and $\| v \|_2 \leq c_2$ for some $c_1,\, c_2 > 0$, where we define $u \coloneqq \dfrac{1+e}{2} l - \dfrac{3-e}{2}m$ and $v \coloneqq l+m$.
    It is easy to check that $l = d_{11} u + d_{12} v$ and $m = d_{21}u + d_{22} v$ for some real numbers $d_{11}, d_{12}, d_{21}, d_{22}$.
    It follows that
    $
        \| l \|_\infty \leq \| l \|_2 \leq d_{11} \| u \|_2 + d_{12} \| v \|_2 \leq d_{11}c_1 + d_{12}c_2
    $
    and
    $
        \| m \|_\infty \leq \| m \|_2 \leq d_{21} \| u \|_2 + d_{22} \| v \|_2 \leq d_{21}c_1 + d_{22}c_2
    $,
    so we have $\| l \|_\infty < R$ and $\| m \|_\infty < R$ whenever $R > \max\{ d_{11}c_1 + d_{12}c_2, d_{21}c_1 + d_{22}c_2\}$. If $(l, m) \in S_2(a)$, then $m = -l$ as $\xi=0$, and the condition $\eta \leq a$ implies that $\| m \|_\infty = \| l \|_\infty \leq \| l \|_2 \leq a / (2\lambda \pi)$.
    % Similarly, if $(l, m) \in S_3(a)$, then $l=\dfrac{3-e}{1+e}m$ as $\eta = 0$, and the condition $\xi \leq a$ implies that $\| m \|_\infty =\dfrac{1+e}{3-e} \| l \|_\infty \leq \| l \|_2 \lesssim a$.
    A similar proof holds for $S_3$.
\end{proof}

\begin{proof}[Proof of Lemma \ref{lemma:decay of the kernel modes}]
    Note that $(l, m) = (0, 0)$ if and only if $\eta = \xi = 0$ and that
    \begin{align}\label{lemma_decay_of_kernel_modes__boundedness_of_indices_1}
        \inf \left\{ \xi: (l, m) \in \mathbb{Z}^d \times \mathbb{Z}^d, \xi \neq 0 \right\}
        &=
        \frac{1+e}{2}\lambda\pi > 0.
    \end{align}
    Also, from the assumption that $e$ is rational, note that
    \begin{align}
    \label{lemma_decay_of_kernel_modes__boundedness_of_indices_2}
        \inf \left\{ \eta: (l, m) \in \mathbb{Z}^d \times \mathbb{Z}^d, \eta \neq 0 \right\}
        &>
        0.
    \end{align}

    %%%%%
    Assume $\eta, \xi > 0$.
    By Lemma~\ref{lemma:appendix_estimation_bessel_0} and the definition of $\mathrm{sinc}$, we have
    \begin{align*}
        &\left| \hat{G}(l, m) \right|
        \lesssim
        \int_0^1 {
            x^{1+\alpha} |J_0(\eta x) J_0(\xi x)|
        }\, dx
        \lesssim
        \int_0^1 {
            \frac{x^{\alpha}}{\sqrt{\eta\xi}}
        }\, dx
        \leq
        \frac{1}{\sqrt{\eta\xi}}
        \quad
        &(d=2),
        \\
        &\left| \hat{G}(l, m) \right|
        \lesssim
        \int_0^1 {
            x^{2+\alpha} |\mathrm{sinc}(\eta x) \mathrm{sinc}(\xi x)|
        }\, dx
        \lesssim
        \int_0^1 {
            \frac{x^{\alpha}}{\eta\xi}
        }\, dx
        \leq
        \frac{1}{\eta\xi}
        \quad
        &(d=3).
    \end{align*}
    Hence, by (\ref{lemma_decay_of_kernel_modes__boundedness_of_indices_1}) and (\ref{lemma_decay_of_kernel_modes__boundedness_of_indices_2}), there is $A_1 > 0$ for which $\left| \hat{G}(l, m)\right| < \epsilon$ whenever $\eta > A_1$ or $\xi > A_1$.
    Moreover, by Lemma~\ref{lemma:appendix_bounded_subset}, there is $K_1 \in \mathbb{N}$ for which $\eta > A_1$ or $\xi > A_1$ whenever $\| l \|_\infty > K_1$ or $\| m \|_\infty > K_1$. Now, assume $\eta > 0$ and $\xi = 0$.
    Then
    \begin{align*}
        \left| \hat{G}(l, m) \right|
        \lesssim
        \int_0^1 {
            x^{1+\alpha} |J_0(\eta x)|
        }\, dx
        \lesssim
        \int_0^1 {
            \frac{x^{\alpha}}{\sqrt{\eta}}
        }\, dx
        \leq
        \frac{1}{\sqrt{\eta}}
        \quad
        &(d=2),
        \\
        \left| \hat{G}(l, m) \right|
        \lesssim
        \int_0^1 {
            x^{2+\alpha} |\mathrm{sinc}(\eta x)|
        }\, dx
        \lesssim
        \int_0^1 {
            \frac{x^{\alpha}}{\eta}
        }\, dx
        \leq
        \frac{1}{\eta}
        \quad
        &(d=3).
    \end{align*}
    Hence, there is a real number $A_2$ for which $\left| \hat{G}(l, m) \right| < \epsilon$ whenever $\eta > A_2$.
    By Lemma~\ref{lemma:appendix_bounded_subset}, there is $K_2 \in \mathbb{N}$ for which $\eta > A_2$ whenever $\| l \|_\infty > K_2$ (or equivalently, $\| m \|_\infty > K_2$).
    Finally, assume $\eta = 0$ and $\xi > 0$.
    By the same argument as in the previous case, there is $K_3 \in \mathbb{N}$ for which $\left| \hat{G}(l, m) \right| < \epsilon$ whenever $\| l \|_\infty > K_3$ (or equivalently, $\| m \|_\infty > K_3$).
    Therefore, choosing $K \coloneqq \max\{K_1, K_2, K_3\} \in \mathbb{N}$, we have $\left| \hat{G}(l, m) \right| < \epsilon$ whenever $\| l \|_\infty,\,  \| m \|_\infty > K$.
\end{proof}

\reviewerTWO{
\section{Ablation study on $N_{\rm{trun}}$, $M$, and the learning rate}
\label{sec:ablation_study}
This section aims to study the effect of the hyperparameters $N_{\rm{trun}}$, $M$, and the learning rate $\eta$ on the training and inference performance of FourierSpecNet.
Ablation study on $N_{\rm{trun}}$, $M$, and $\eta$ are provided in Appendix B.~{1, 2, 3}, respectively.

All models were trained to approximate the Boltzmann collision operator with the elasticity parameter set to $e=0.2$.
Training was performed for $500,000$ epochs using an NVIDIA A100 GPU on PyTorch 2.5.1.
After training, the models were used to solve the inelastic Boltzmann equation with the initial condition in (\ref{eq:init_inela_2}).
Note that the target kinetic energy in each of the following experiments is evaluated using the numerical solution for the input resolution $N$ of $256$, which is computed using the fast spectral method.

\subsection{Ablation study on $N_{\rm{trun}}$}
\label{subsec:ablation_study__degree}
    To examine the influence of $N_{\mathrm{trun}}$, we conducted an ablation study by fixing $M = 2$ and the learning rate at $10^{-2}$, while varying $N_{\mathrm{trun}} \in \{4, 8, 16, 32\}$.

    % Trian/validation/test curves
    \begin{figure}[h]
        \includegraphics[width=\textwidth]{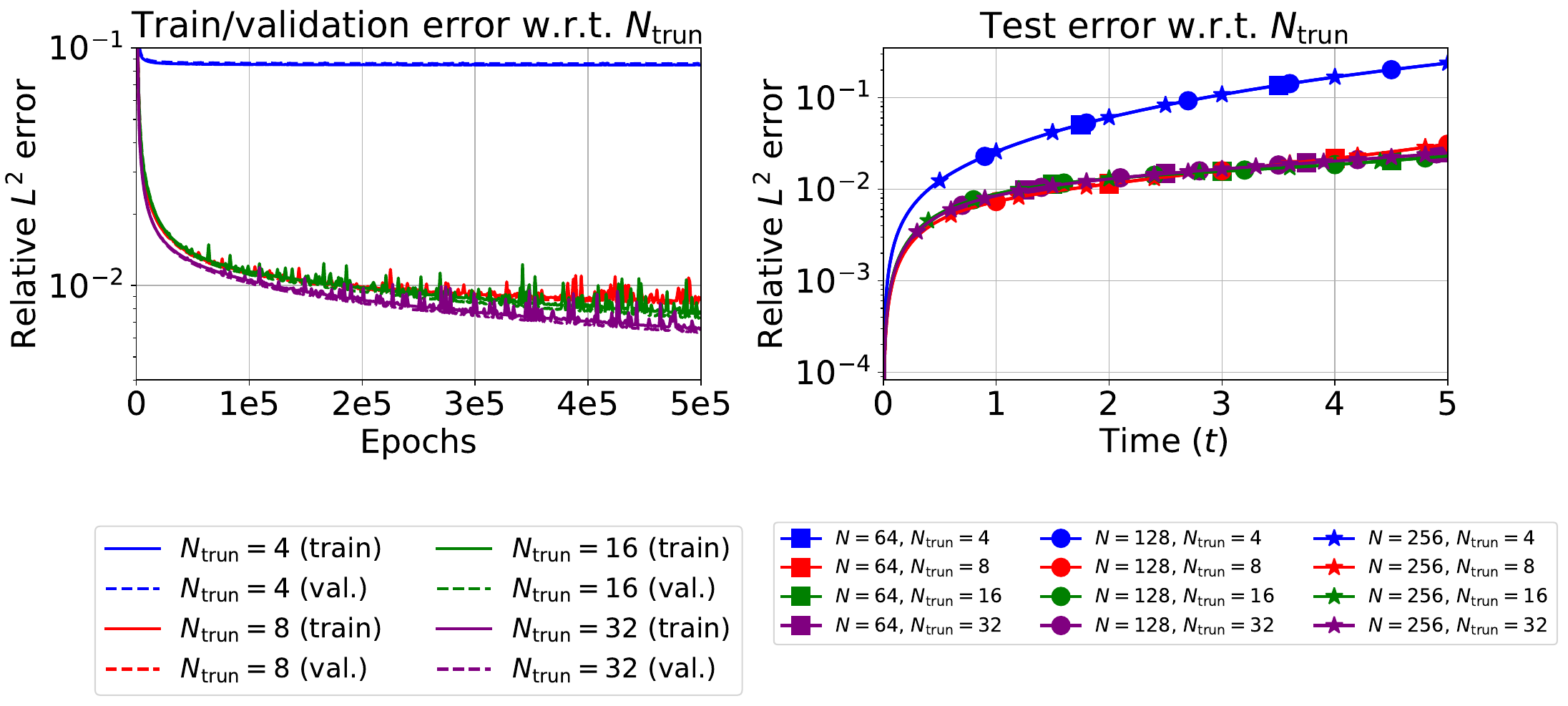}
        \caption{
            (Left)
            Evolution of training and validation errors during model training for different values of $N_{\rm{trun}}$.
            (Right)
            Simulation results obtained from the trained models for different values of $N_{\rm{trun}}$.
            The simulation is performed with the initial condition $f^{(2)}_{\rm{init}}$ using the RK3 scheme with a time step of $\Delta t = 0.01$.
        }
        \label{fig:ablation__degree__train_valid_test}
    \end{figure}
    % Training times
    \begin{table}[h]
        \centering
        \begin{tabular}{c|cccc}
\hline
$N_{\rm{trun}}$ & 4 & 8 & 16 & 32\\
\hline
\# of parameters & 768 & 3072 & 12288 & 49152\\
\hline
Training time & 04h 16m & 04h 19m & 04h 22m & 04h 21m\\
\hline
\end{tabular}
        
        \caption{The training times of FourierSpecNet for different values of $N_{\rm{trun}}$.}
        \label{tab:training_time__degree}
    \end{table}
    % Inference times
    \begin{table}[h]
        \centering
        \begin{tabular}{c|cccc}
\hline
\diagbox{$N$}{$N_{\rm{trun}}$} & 4 & 8 & 16 & 32\\
\hline
64  & 0.47s & 0.47s & 0.46s & 0.46s\\
\hline
128 & 0.49s & 0.49s & 0.48s & 0.48s\\
\hline
256 & 0.52s & 0.51s & 0.51s & 0.52s\\
\hline
\end{tabular}
% \begin{tabular}{c|cccc}
% \hline
% \diagbox{$N$}{$N_{\rm{trun}}$} & 4 & 8 & 16 & 32\\
% \hline
% 64  & 0.4654s & 0.4675s & 0.4645s & 0.4592s\\
% \hline
% 128 & 0.4882s & 0.4858s & 0.4788s & 0.4809s\\
% \hline
% 256 & 0.5152s & 0.5138s & 0.5146s & 0.5169s\\
% \hline
% \end{tabular}
        
        \caption{The inference times of FourierSpecNet for different values of $N_{\rm{trun}}$ and input resolution $N$, measured in seconds.}
        \label{tab:inference_time__degree}
    \end{table}
    % Physical quantities
    \begin{figure}[h]
        \includegraphics[width=\textwidth]{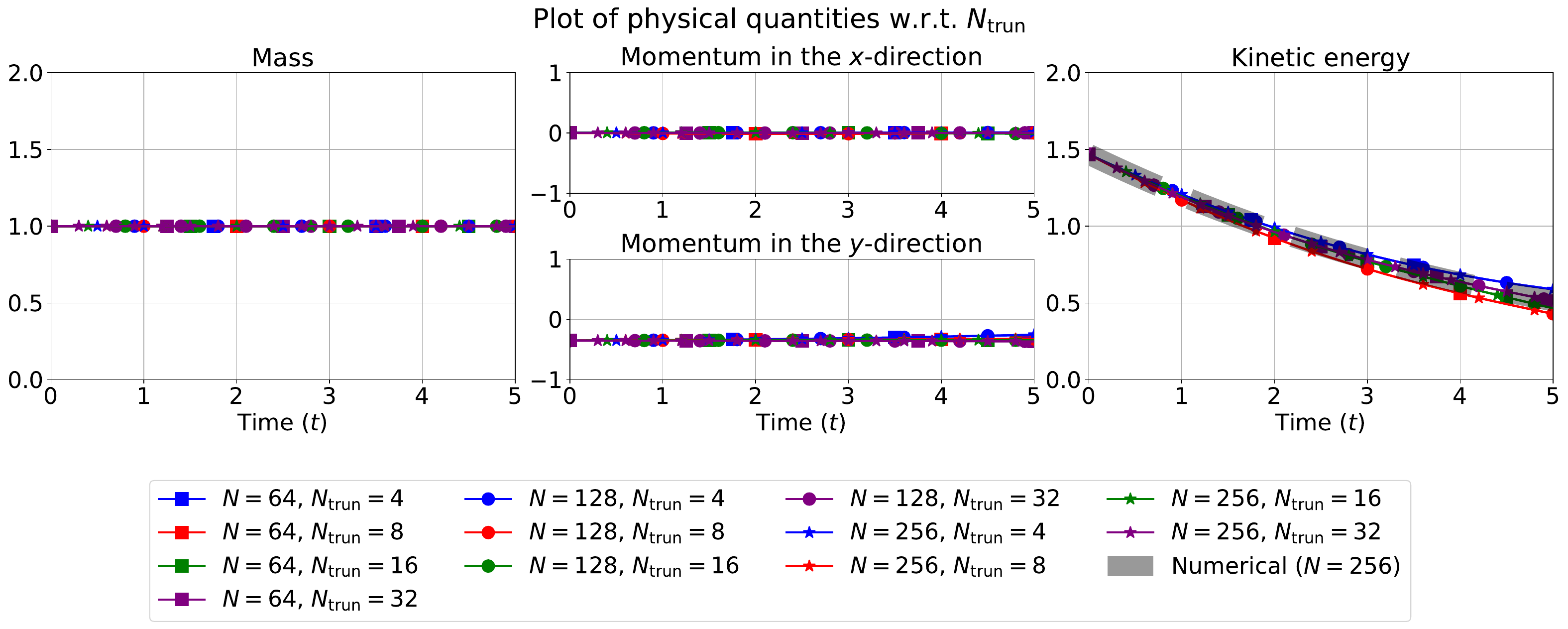}
        \caption{
            Plot of physical quantities of the simulation result for various values of $N_{\rm{trun}}$.
            (Left)
            The plot of the mass.
            (Middle above)
            The plot of the $x$-direction of the momentum.
            (Middle below)
            The plot of the $y$-direction of the momentum.
            (Right)
            The plot of the kinetic energy.
            Since the collision is inelastic, the kinetic energy decays in time $t$.
        }
        \label{fig:ablation__degree__physical_quantities}
    \end{figure}
    
    Figure~\ref{fig:ablation__degree__train_valid_test} shows the training histories (left panel) of the models, and the relative $L^2$ errors of the predicted solutions (right panel) corresponding to $N_{\rm{trun}} \in \{4, 8, 16, 32\}$.
    As observed in the left panel, FourierSpecNet with $N_{\rm{trun}} \geq 8$ learns the collision operator, with a relative $L^2$ error below $10^{-2}$.
    In contrast, the model with $N_{\rm{trun}} = 4$ does not successfully learn the collision operator, and its relative $L^2$ error saturates around $8 \times 10^{-2}$.
    The right panel demonstrates that during inference, FourierSpecNet achieves a relative $L^2$ error below $3.2\%$ for $N_{\rm{trun}} \geq 8$, whereas the case with $N_{\rm{trun}} = 4$ yields a relative error of $23.9\%$ at the terminal time $T=5$.
    The corresponding training and inference times for different $N_{\rm{trun}}$ values are summarized in Table~\ref{tab:training_time__degree} and Table~\ref{tab:inference_time__degree}, respectively.
    Table~\ref{tab:training_time__degree} shows that the maximum difference of the training times is about $6$ minutes, which is the difference of the training times for $N_{\rm{trun}} = 4$ and $N_{\rm{trun}} = 16$.
    This difference accounts for less than $2.4\%$ of the shortest training time ($N_{\rm{trun}}=4$), implying that there is no significant difference between the training times across $N_{\rm{trun}}$ values.
    Table~\ref{tab:inference_time__degree} shows that the inference times of the models trained for differnent values of $N_{\rm{trun}}$, and it turns out that the inference times of the models are not significantly different for each input resolution $N$.
    
    Beyond the training and inference accuracy, we also examined the conservation of mass and momentum and the decay of kinetic energy during inference.
    Figure \ref{fig:ablation__degree__physical_quantities} illustrates these physical quantities for $N_{\rm{trun}} \in \{4, 8, 16, 32\}$ and the input resolutions $N \in \{64, 128, 256\}$.
    The left and middle panels confirm that FourierSpecNet preserves mass and momentum across all tested configurations.
    The right panel shows that the kinetic energy of the predicted solutions monotonically decays over time, consistent with the inelastic collision assumption.
    These results further verify that FourierSpecNet produces physically consistent and reliable solutions.

    %%%%%
    From the above results, we conclude that FourierSpecNet successfully learns the collision operator for $N_{\rm{trun}} = 8$, also minimizing the number of trainable parameters without loss of performance.
    
    % From the above results, we conclude that FourierSpecNet successfully learns the collision operator for $N_{\rm{trun}} = 8$ and $M = 2$ within the shortest time.
    % Although the trained FourierSpecNet with $M=5$ performs better than the trained FourierSpecNet with $M=2$, as FourierSpecNet learns the collision operator about $34.0\%$ faster for $M=2$ than for $M=5$ without showing significant difference in the inference error, the manuscript chooses $N_{\rm{trun}} = 8$ and $M = 2$, which also minimizes the number of trainable parameters without loss of performance.

\subsection{Ablation study on $M$}
\label{subsec:ablation_study__rank}
    Similarly, to investigate the effect of $M$, we performed another ablation study by fixing $N_{\mathrm{trun}} = 8$ and the learning rate to $10^{-2}$, while varying $M \in \{2, 5, 8, 16\}$.    
    Figure~\ref{fig:ablation__rank__train_valid_test} presents the training histories and inference results for different values of $M$.

    % Train/validation/test curves
    \begin{figure}[h]
        \includegraphics[width=\textwidth]{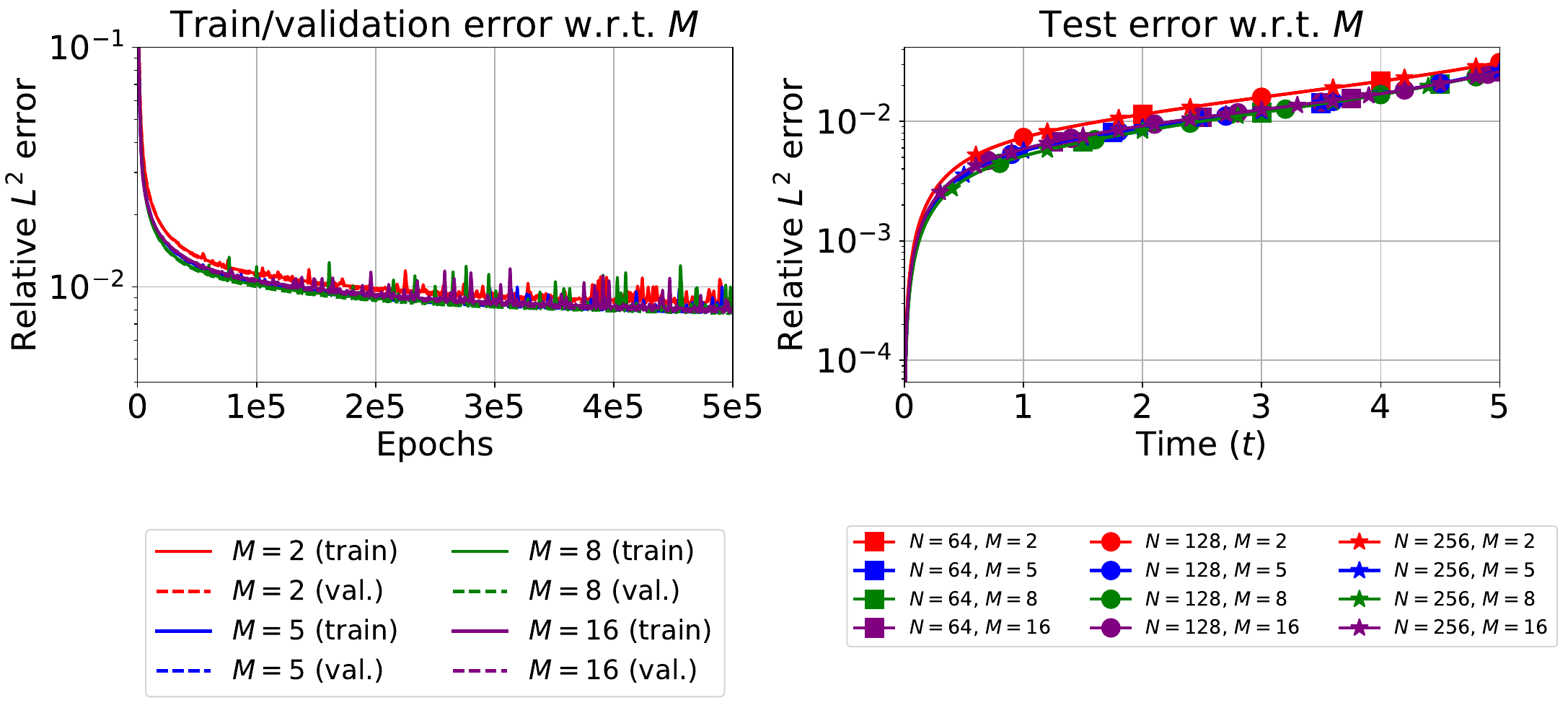}
        \caption{
            (Left)
            The plot of the train error and the validation error throughout the training models for different values of $M$.
            (Right)
            The plot of the simulation result of the trained models for different values of $M$.
        }
        \label{fig:ablation__rank__train_valid_test}
    \end{figure}
    % Training times
    \begin{table}[h]
        \centering
        \begin{tabular}{c|cccc}
\hline
$M$ & 2 & 5 & 8 & 16\\
\hline
\# of parameters & 3072 & 7680 & 12288 & 24576\\
\hline
Training time & 04h 19m & 05h 47m & 07h 17m & 11h 17m\\
\hline
\end{tabular}
        \caption{The training times of FourierSpecNet for different values of $M$.}
        \label{tab:training_time__rank}
    \end{table}
    % Inference times
    \begin{table}[h]
        \centering
        \begin{tabular}{c|cccc}
\hline
\diagbox{$N$}{$M$} & 2 & 5 & 8 & 16\\
\hline
64  & 0.47s & 0.46s & 0.46s & 0.46s\\
\hline
128 & 0.48s & 0.48s & 0.48s & 0.48s\\
\hline
256 & 0.52s & 0.52s & 0.52s & 0.66s\\
\hline
\end{tabular}
% \begin{tabular}{c|cccc}
% \hline
% \diagbox{$N$}{$M$} & 2 & 5 & 8 & 16\\
% \hline
% 64  & 0.4681s & 0.4620s & 0.4619s & 0.4634s\\
% \hline
% 128 & 0.4770s & 0.4825s & 0.4834s & 0.4758s\\
% \hline
% 256 & 0.5160s & 0.5179s & 0.5163s & 0.6635s\\
% \hline
% \end{tabular}
        \caption{The inference times of FourierSpecNet for various values of $M$ and input resolution $N$.}
        \label{tab:inference_time__rank}
    \end{table}
    % Physical quantities
    \begin{figure}[h]
        \includegraphics[width=\textwidth]{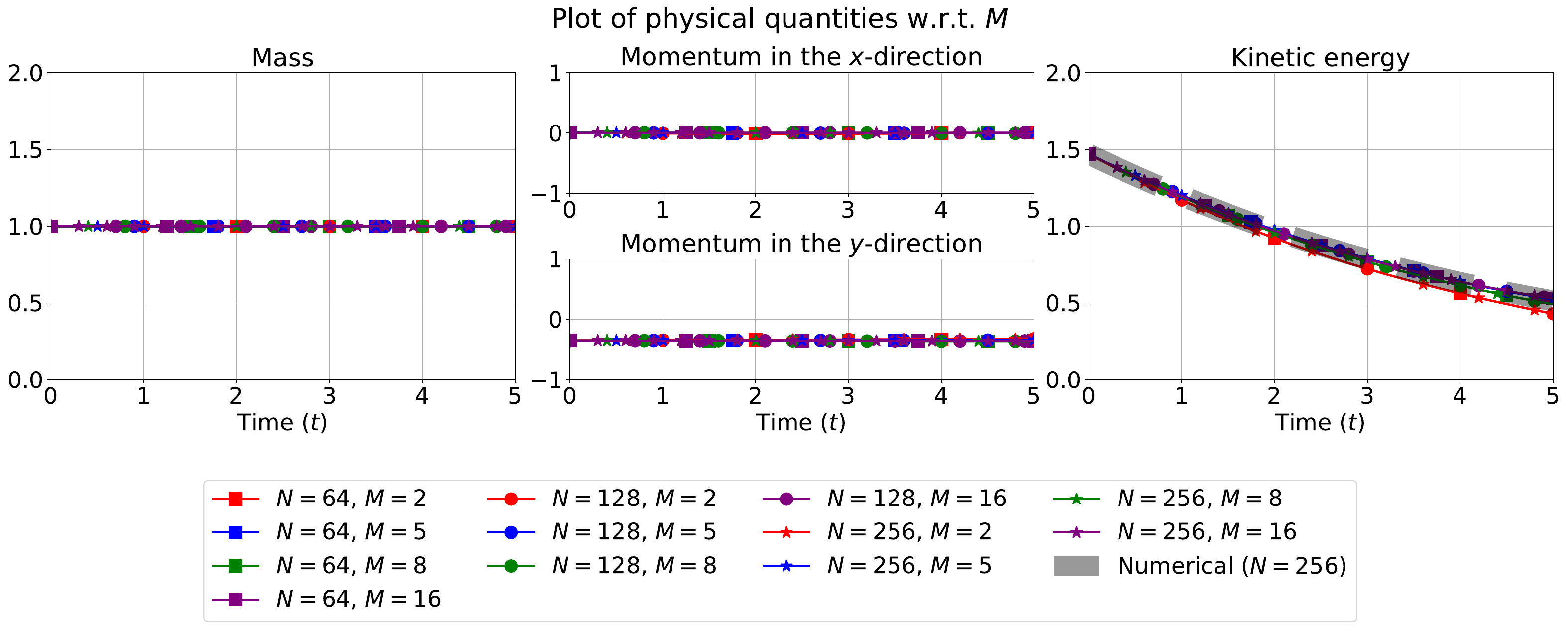}
        \caption{
            Plot of physical quantities of the simulation result for various values of $N_{\rm{trun}}$.
            (Left)
            The plot of the mass.
            (Middle above)
            The plot of the $x$-direction of the momentum.
            (Middle below)
            The plot of the $y$-direction of the momentum.
            (Right)
            The plot of the kinetic energy.
            Since the collision is inelastic, the kinetic energy decays in time $t$.
        }
        \label{fig:ablation__rank__physical_quantities}
    \end{figure}

    As shown in the left panel, FourierSpecNet converges faster and achieves lower training and validation errors as $M$ increases up to $M = 8$, whereas the error for $M = 16$ is slightly larger than that for $M = 5$.
    The right panel illustrates the relative $L^2$ error of the predicted solutions.
    Although the inference accuracy improves when $M \geq 5$ compared to $M = 2$, there is no significant difference in accuracy among models with $M \geq 5$.
    This observation indicates that the low-rank structure of FourierSpecNet can effectively capture the essential dynamics of the collision operator with a moderate number of rank components.
    The training times for different values of $M$ are summarized in Table~\ref{tab:training_time__rank}.
    As expected, the training time increases with $M$ due to the larger number of trainable parameters.
    Considering both inference accuracy and training cost, the configuration with $M = 5$ can be regarded as the most efficient choice for FourierSpecNet.

    In addition, we evaluated the conservation of physical quantities during inference.
    Figure~\ref{fig:ablation__rank__physical_quantities} shows the evolution of mass, momentum, and kinetic energy for different values of $M$.
    FourierSpecNet successfully preserves the conservation of mass and momentum regardless of $M$, and consistently reproduces the expected decay of kinetic energy over time due to inelastic collisions.
    These results confirm that FourierSpecNet yields physically consistent solutions even when the model complexity (rank $M$) varies.

    From the above results, we conclude that FourierSpecNet successfully learns the collision operator for $M = 2$ within the shortest time.
    Although the trained FourierSpecNet with $M=5$ performs better than the trained FourierSpecNet with $M=2$, FourierSpecNet learns the collision operator about $34.0\%$ faster for $M=2$ than for $M=5$ without showing a significant difference in the inference error. Therefore, we choose $N_{\rm{trun}} = 8$ and $M = 2$, which also minimizes the number of trainable parameters without loss of performance.

\subsection{Ablation study on $\eta$}
\label{subsec:ablation_study__learning_rate}
    To examine the effect of the learning rate, we performed an ablation study with $N_{\mathrm{trun}} = 8$ and $M = 2$, while varying the learning rate $\eta \in \{10^{-2}, 5 \times 10^{-3}, 2 \times 10^{-3}, 10^{-3}, 5 \times 10^{-4}, 2 \times 10^{-4}, 10^{-4}\}$.

    % Train/validation/test curves
    \begin{figure}[h]
        \includegraphics[width=\textwidth]{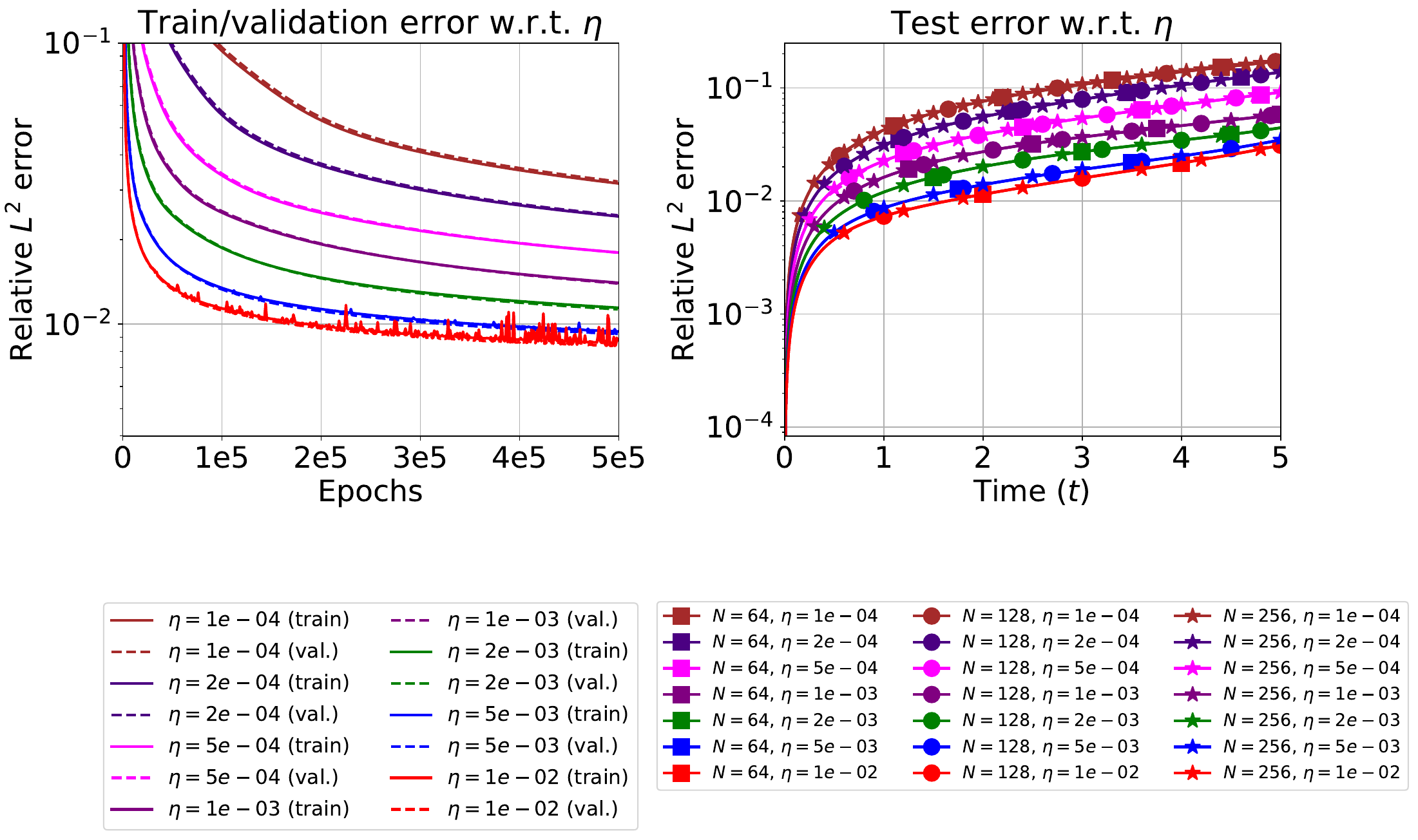}
        \caption{
            (Left)
            The plot of the train error and the validation error throughout the training models for different values of $\eta$.
            (Right)
            The plot of the simulation result of the trained models for different values of $\eta$.
        }
        \label{fig:ablation__learning_rate__train_valid_test}
    \end{figure}
    % Training times (Manual resizing)
    \begin{table}[h]
        \centering
        \resizebox{\linewidth}{!}{
            \begin{tabular}{c|ccccccc}
\hline
$\eta$ & $10^{-2}$ & $5 \times 10^{-3}$ & $2 \times 10^{-3}$ & $10^{-3}$ & $5 \times 10^{-4}$ & $2 \times10^{-4}$ & $10^{-4}$\\
\hline
Training time & 04h 19m & 04h 13m & 04h 16m & 04h 09m & 04h 16m & 04h 21m & 03h 58m\\
\hline
\end{tabular}
        }
        \caption{The training times of FourierSpecNet for different values of learning rate $\eta$.}
        \label{tab:training_time__learning_rate}
    \end{table}
    % Inference times
    \begin{table}[h]
        \centering
        \begin{tabular}{c|ccccccc}
\hline
\diagbox{$N$}{$\eta$} & $10^{-2}$ & $5 \times 10^{-3}$ & $2 \times 10^{-3}$ & $10^{-3}$ & $5 \times 10^{-4}$ & $2 \times 10^{-4}$ & $10^{-4}$\\
\hline
64  & 0.46s & 0.45s & 0.45s & 0.46s & 0.47s & 0.46s & 0.46s\\
\hline
128 & 0.48s & 0.49s & 0.49s & 0.47s & 0.48s & 0.48s & 0.47s\\
\hline
256 & 0.51s & 0.52s & 0.52s & 0.51s & 0.52s & 0.52s & 0.51s\\
\hline
\end{tabular}
% \begin{tabular}{c|ccccccc}
% \hline
% \diagbox{$N$}{$\eta$} & $10^{-2}$ & $5 \times 10^{-3}$ & $12 \times 10^{-3}$ & $10^{-3}$ & $5 \times 10^{-4}$ & $2 \times 10^{-4}$ & $10^{-4}$\\
% \hline
% 64  & 0.4624s & 0.4522s & 0.4544s & 0.4559s & 0.4691s & 0.4617s & 0.4602s\\
% \hline
% 128 & 0.4828s & 0.4859s & 0.4858s & 0.4710s & 0.4830s & 0.4808s & 0.4709s\\
% \hline
% 256 & 0.5097s & 0.5180s & 0.5221s & 0.5091s & 0.5194s & 0.5179s & 0.5146s\\
% \hline
% \end{tabular}
        \caption{The inference times of FourierSpecNet for different values of learning rate $\eta$.}
        \label{tab:inference_time__learning_rate}
    \end{table}
    % Physical quantities
    \begin{figure}[h]
        \includegraphics[width=\textwidth]{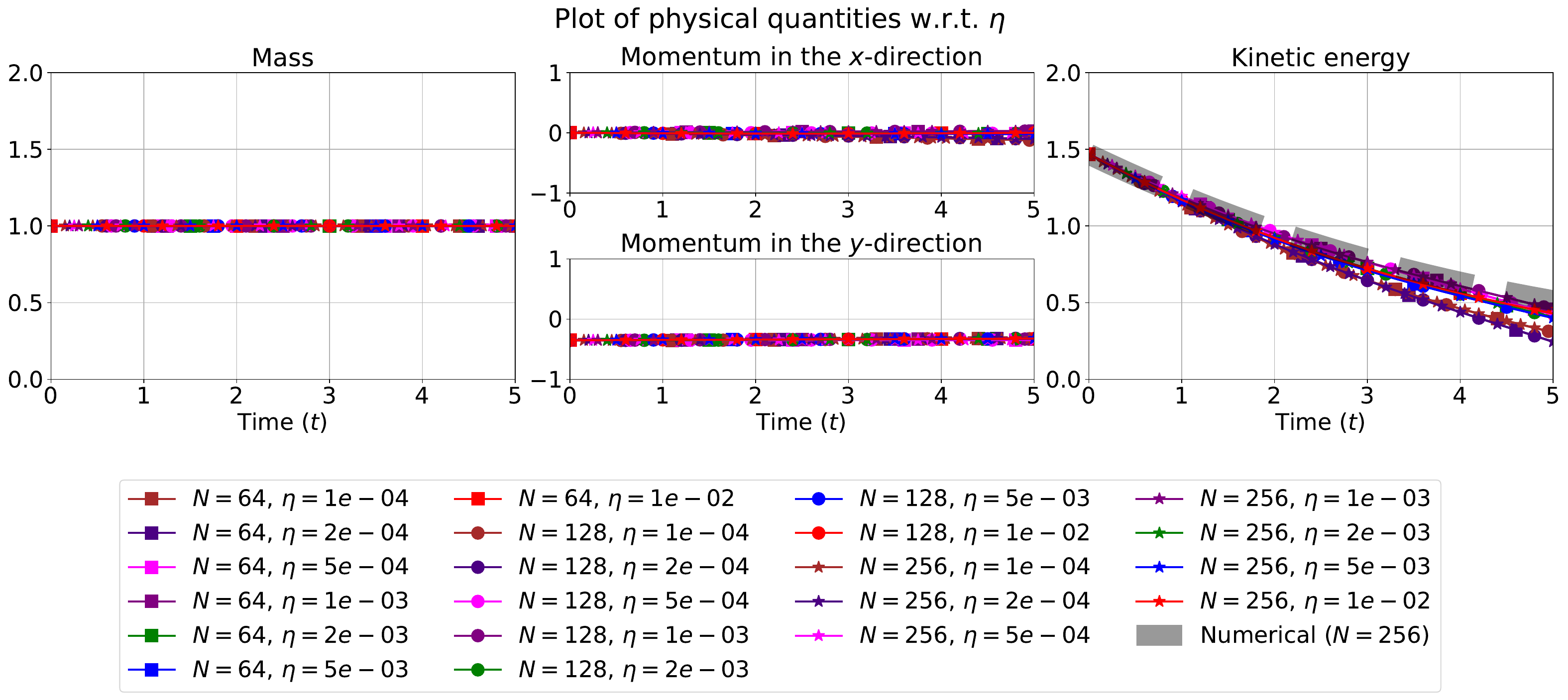}
        \caption{
            Plot of physical quantities of the simulation result for various values of $\eta$.
            (Left)
            The plot of the mass.
            (Middle above)
            The plot of the $x$-direction of the momentum.
            (Middle below)
            The plot of the $y$-direction of the momentum.
            (Right)
            The plot of the kinetic energy.
            Since the collision is inelastic, the kinetic energy decays in time $t$.
        }
        \label{fig:ablation__learning_rate__physical_quantities}
    \end{figure}
    Figure~\ref{fig:ablation__learning_rate__train_valid_test} (left) shows the training and validation losses during the learning process.
    As $\eta$ decreases from $10^{-2}$ to $10^{-4}$, the training becomes noticeably slower, and the model converges to a higher loss.
    In the right panel of Figure~\ref{fig:ablation__learning_rate__train_valid_test}, the relative $L^2$ errors of the inferred solutions are plotted, showing that smaller learning rates result in higher errors.
    Table~\ref{tab:training_time__learning_rate} reports that the total training time does not vary significantly with the learning rate, and Table~\ref{tab:inference_time__learning_rate} shows that the inference time also does not vary significantly with the learning rate.

    We further examined the physical consistency of the inferred solutions under different learning rates.
    As shown in Figure~\ref{fig:ablation__learning_rate__physical_quantities}, the mass is well conserved for all learning rates, while the momentum conservation deteriorates for $\eta = 10^{-4}$.
    The kinetic energy decays over time in all cases, which is consistent with the inelastic nature of the collision.
    These results demonstrate that the learned models capture physically meaningful behavior across different learning rates.

    Considering the training and inference errors, the physical consistency, and the computational cost, we conclude that $\eta = 10^{-2}$ provides the best balance between convergence speed and stability.
}

\section*{Acknowledgement}
This research was supported by the
Chung-Ang University Research Grants in 2024.
Gwang Jae Jung, Byungchan Lim and Hyung Ju Hwang were supported by the National Research Foundation of Korea (NRF) grant funded by the Korea government (MSIT) (No. RS-2023-00219980).

%%%%%%%%%%%%%%%%%%%%%%%%%%%%%%%%%%%%%%%%%%%%%%%%%%%%%%%%%%%%%%%%%%%%%%%%%%%%%%%%%%

% these commends below are for a bibtex bibliography THEY CHANGE THE STYLE OF THE BIBLIOGRAPHY
%\bibliographystyle{plain}
%\bibliographystyle{abbrv}
%\bibliographystyle{acm}
%\bibliographystyle{alpha}
%\bibliographystyle{apalike}
%\bibliographystyle{ieeetr}
%\bibliographystyle{siam}
%\bibliographystyle{unsrt}
%\bibliographystyle{plainnat}
%\bibliographystyle{plainurl}
%\bibliographystyle{abbrvnat}
%\bibliographystyle{unsrtnat}
%\bibliographystyle{amsalpha.bst}  % this one is cool with amsrefs
%\bibliographystyle{amsplain.bst}
%% the ones below need to be called with AMSREFS package
%\bibliographystyle{amsrn.bst}  %this one is the default if no style is called
%\bibliographystyle{amsru.bst}
%\bibliographystyle{amsra.bst}
%\bibliographystyle{amsry.bst}
%\bibliographystyle{amsrs.bst}
%\bibliographystyle{amsxport.bst}
%%% the above STYLE FILES ARE IN MOST LATEX INSTALLATIONS

%%%%% the bibliography styles below support arXiv eprint links
%\bibliographystyle{hplain}. % THIS STYLE FILE INCLUDES EPRINT INFORMATION IN THE BIB
% hplain was downloaded from https://arxiv.org/help/hypertex/bibstyles
%\bibliographystyle{habbrv}  % INCLUDES DOI AND ARXIV LINKS AND USES NUMBER REFERENCES
%\bibliographystyle{halpha}  %INCLUDES DOI AND ARXIV LINKS AND USES NAME REFERCES
% the last two were downloaded from http://www.math.cmu.edu/~gautam/sj/blog/20171114-bibtex-doi.html

\bibliographystyle{amsplaindoi} %AMS references format with number references, and clickable URL links
\bibliography{arxiv_bibliograpy}

%%% input the BBL file when done using BibTeX \left(change the name to be the document name\right)
%\input{RBELinfty.bbl}

%%%%%%%%%%%%%%%%%%%%%%%%%%%%%%%%%%%%%%%%%%%%%%%%%%%%%%%%%%%%%%%%%%%%%%%%%%%%%%%%%%

%%% copy and paste the BBL file below here when you are ready to upload, and comment the above bibtex library

%%\newpage

\end{document}